\documentclass[10pt,journal,compsoc]{IEEEtran}
\usepackage[nocompress]{cite}
\usepackage{cite}
\usepackage{makeidx}
\usepackage{graphicx}
\usepackage{multicol}
\usepackage[bottom]{footmisc} 
\usepackage{amsmath, amsfonts, amssymb, amsthm, bm}
\usepackage{algorithmic}
\usepackage{array}
\usepackage{subcaption}
\usepackage{textcomp}
\usepackage{stfloats}
\usepackage{url}
\usepackage{verbatim}
\usepackage{makecell}
\usepackage{hyperref}
\usepackage{booktabs}
\usepackage{multirow}
\usepackage{tablefootnote}
\usepackage{xcolor}
\usepackage{float}
\usepackage{lipsum}
\usepackage[english]{babel}
\usepackage[nameinlink]{cleveref}
\crefformat{footnote}{#2\footnotemark[#1]#3}
\newcommand{\rev}[1]{{\color{black}#1}}

\begin{document}

\title{Convolutional and Deep Learning based techniques for Time Series Ordinal Classification}

\author{Rafael Ayllón-Gavilán,
        David Guijo-Rubio$^*$, \textit{Member, IEEE},
        Pedro Antonio Gutiérrez, \textit{Senior Member, IEEE},
        Anthony Bagnall,
        César Hervás-Martínez, \textit{Senior Member, IEEE},
\IEEEcompsocitemizethanks{

\IEEEcompsocthanksitem $^*$: Corresponding author.\protect\\
\IEEEcompsocthanksitem R. Ayllón-Gavilán is with the Dept. of Clinical-Epidemiological Research in Primary Care, IMIBIC, Spain. E-mail: rafael.ayllon@imibic.org.\protect\\
\IEEEcompsocthanksitem D. Guijo-Rubio, P.A. Gutiérrez, and C. Hervás-Martínez are with the Dept. of Computer Science and Numerical Analysis, University of Cordoba, Spain. E-mail: \{dguijo,pagutierrez,chervas\}@uco.es.\protect\\
\IEEEcompsocthanksitem A. Bagnall is with the Dept. of Electronics and Computer Science, University of Southampton, United Kingdom. E-mail: A.J.Bagnall@soton.ac.uk.}
\thanks{Manuscript received June 18, 2023; revised November 4, 2024.}}

\markboth{IEEE Transactions on Cybernetics}%
{Ayllón-Gavilán \MakeLowercase{\textit{et al.}}: Convolutional and Deep Learning based techniques for TSOC}

\IEEEtitleabstractindextext{%
\begin{abstract}
Time Series Classification (TSC) covers the supervised learning problem where input data is provided in the form of series of values observed through repeated measurements over time, and whose objective is to predict the category to which they belong. When the class values are ordinal, classifiers that take this into account can perform better than nominal classifiers. Time Series Ordinal Classification (TSOC) is the field bridging this gap, yet unexplored in the literature. There are a wide range of time series problems showing an ordered label structure, and TSC techniques that ignore the order relationship discard useful information. Hence, this paper presents the first benchmarking of TSOC methodologies, exploiting the ordering of the target labels to boost the performance of current TSC state-of-the-art. Both convolutional- and deep learning-based methodologies (among the best performing alternatives for nominal TSC) are adapted for TSOC. For the experiments, a selection of $29$ ordinal problems has been made. In this way, this paper contributes to the establishment of the state-of-the-art in TSOC. The results obtained by ordinal versions are found to be significantly better than current nominal TSC techniques in terms of ordinal performance metrics, outlining the importance of considering the ordering of the labels when dealing with this kind of problems.
\end{abstract}

}

\maketitle

\IEEEdisplaynontitleabstractindextext

\IEEEpeerreviewmaketitle

\section{Introduction}\label{cap:intro}

A time series is an ordered sequence of values. This type of data is found in a wide variety of domains, such as medicine \cite{time_series_for_medicine_3}, financial analysis \cite{time_series_for_finance_2}, \rev{and} agriculture \cite{time_series_for_agriculture_2}. For instance, electrocardiogram signals \cite{atrial_fibrillation} are ordered by time, and spectrograms \cite{ethanol_level} are ordered by frequency. \rev{There are numerous tasks applicable to time series, such as forecasting the next value \cite{cybernetics_forecasting,jin2024survey}, clustering time series into groups without class information \cite{cybernetics_guiyo,holder2024review}, performing extrinsic regression on time series \cite{mohammadi2024deep}, and detecting anomalies within the data \cite{zamanzadeh2024deep}, among others. However, Time Series Classification (TSC) is the most popular Machine Learning (ML) tasks, with hundreds of different approaches proposed in the literature \cite{time_series_bakeoff, bakeoff_redux}. For instance, \cite{jastrzebska2023fuzzy} proposed a fuzzy-driven methodology, while \cite{mei2016learning} focused on classifying time series based on the distance between them. More recent approaches aim to classify time series using subsequences that capture the characteristics of the entire series \cite{wan2024memory}, exploring the use of transformers applied to time series \cite{foumani2024improving}, and reservoir models based on spiking neural P systems \cite{peng2024reservoir}. Moreover, other time series related domains are time series segmentation \cite{carmona2024optimal} and the discovery of motifs \cite{schafer2022motiflets,schafer2024discovering}, with an increasing interest in the last years.}

TSC involves predicting a discrete output variable for a given time series. Depending on the number of variables observed at each time point, time series are univariate (only one channel) or multivariate (two or more channels) \cite{tsc_explanation,garcia2014structural}. The publication of the TSC archive\footnote{\url{https://timeseriesclassification.com/}} \cite{ucr} allowed the development of effective methods for TSC. This archive provides a heterogeneous problem set that facilitates objective comparisons of new algorithms: a recent bake off study  compared $33$ TSC algorithms proposed in the last five years using the UCR archive\cite{bakeoff_redux}.

While there has been significant progress in the algorithmic development for TSC, almost no attention has been paid to Time Series Ordinal Classification (TSOC). We aim to address this absence. Some problems in the UCR archive are ordinal in nature, \rev{i.e. labels associated with samples follow an ordinal relationship.} Up to now, these problems have been tackled by nominal TSC methods, which can limit the learning process: nominal methods generally require more data or iterations to achieve the same performance as an ordinal classifier \cite{Harrington2003}.

We can define ordinal classification (also known as ordinal regression) as a classification problem where the output labels exhibit a natural ordering. This characteristic can be found in many and varied domains \rev{such as} human age prediction \cite{cybernetics_ordinal_human_age_prediction}, climatological applications \cite{ordinal_for_army} \rev{and} medical research \cite{ordinal_for_covid_china}. \rev{This research takes} advantage of information related to the ordering of the categorical labels to increase the performance of the models being applied. The most prominent type of models in the literature on ordinal classification are the so-called threshold-based models. In these methodologies the existence of a real-valued variable underlying the ordinal response is assumed. Hence, the training process focuses on modelling the real variable and learning the optimal thresholds determining which intervals corresponds to each ordinal label. The Cumulative Link Model (CLM) approach belongs to this family of threshold-based models, and works with cumulative probabilities of the input belonging to a certain class or classes \rev{lower} in the ordinal scale.

An example of ordinal classification can be seen in \cite{ordinal_for_covid_china}, where a set of patients infected with the SARS-CoV-2 coronavirus is studied and grouped into three levels of illness severity: moderate, severe, and critical. These labels follow an ordinal relationship, in that a critical patient is sicker than a moderate or a severe patient. The main characteristic of an ordinal variable, \rev{using} this example, is that it is worse to misclassify a critical patient as moderate, than it is to misclassify them as severe. The magnitude of the error should be higher in the first case.

As stated in \cite{anderson_ordinal_variables}, two types of ordinal problems can be distinguished in the literature: 1) grouped continuous variables, and 2) assessed ordered categorical variables. The former involves an underlying continuous variable that is divided into different categories through a discretisation procedure. In contrast, the latter does not involve a continuous variable, and instead, a domain expert assigns labels to patterns, establishing the ordering based on their judgement. An example of the first type is predicting the price of a computer for the next week in Euros in several categories (e.g., 0-1000, 1001-2000, 2001-3000, and so on). In this case, the categories are more objective as they are not influenced by human opinion. On the other hand, an example of the second type is assessing the severity of an investment risk based on its trajectory (e.g., none, mild, moderate, severe, and critical). In this scenario, \rev{the categories} may be subjective, as different experts in the domain may have varying opinions and could assign the same risk to different trajectories. In this work, our focus lies on the first type of ordinal problems, as all the TSOC problems sourced from various data repositories involve the discretisation of an underlying continuous variable. However, the methods are also applicable to the second type of problems.

\begin{figure}[ht]
	\centering
	\includegraphics[width=0.9\linewidth]{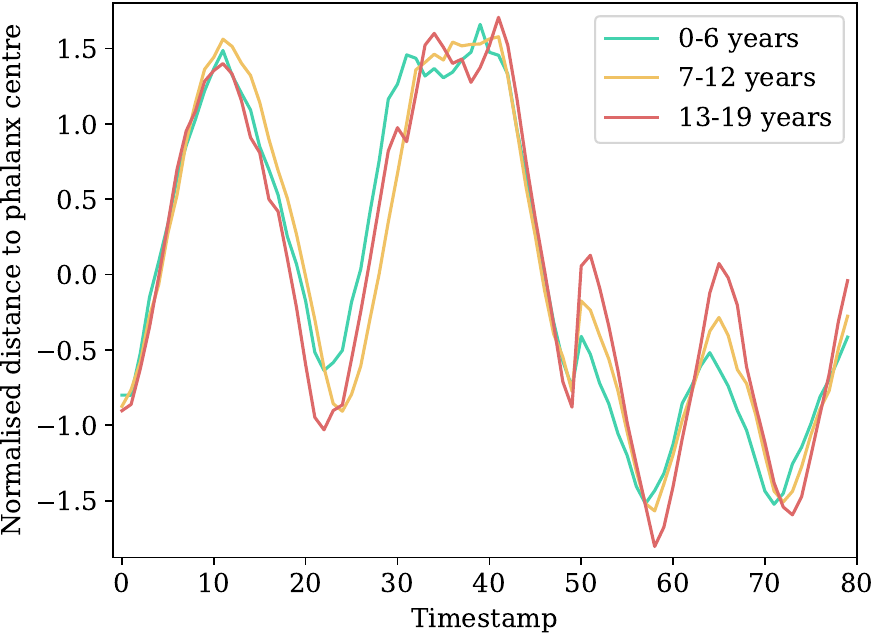}
	\caption{Example of time series extracted from the \textit{DistalPhalanxOAG} dataset. The target ordinal scale represents different age ranges.}
	\label{fig:ordinal_time_series_example}
\end{figure}

\rev{An example of dataset used in this study is the DistalPhalanxOAG dataset \cite{bagnall2014predictive}}. It is specifically designed to assess the effectiveness of detecting hand and bone outlines and determine if they can aid in predicting bone age. The dataset focuses on distal phalanges of the middle finger, and the labels correspond to different age groups: 0-6 years old, 7-12 years old, and 13-19 years old. Figure \ref{fig:ordinal_time_series_example} illustrates three patterns from each of the classes, showcasing the observable ordinality of the labels in certain parts of the time series.

Other domains that could benefit from the development of the TSOC field include cardiology \cite{atrial_fibrillation}, where the objective is to develop methodologies for predicting the spontaneous termination of Atrial Fibrillations (AFs). \rev{Here, the input time series consists of two-channel electrocardiogram recordings from patients affected by AFs, and the aim is to predict whether the AF will terminate within one second, one hour, or more than one hour}. In the field of spirit authentication, researchers in \cite{ethanol_level} utilised non-invasive near infrared spectroscopy time series to classify alcohol content into one of the following levels: E35, E38, E40, and E45, which clearly demonstrate an order relationship. Furthermore, another area is in stock market prediction \cite{shabani2023augmented}. The authors presented a methodology for predicting the direction (decreasing, stationary, increasing) of the mid-price for various prediction horizons. In this scenario, there exists a natural order between the labels, making it suitable for TSOC analysis.

\rev{Our contributions can be listed as follows: 1) the development of seven novel TSOC techniques, focusing on the state-of-the-art approaches in TSC. 2) The identification of $29$ ordinal time series problems from different data sources (TSC, TSER, and development of new TSOC datasets), and introducing the University of Córdoba (UCO) TSOC repository. 3) A benchmarking study of all these seven techniques over the whole set of $29$ ordinal datasets. 4) The proposed approaches have been compared against their nominal counterpart (TSC approaches), the state-of-the-art HIVE-COTE2 (HC2), and standard ML approaches. The results demonstrate that TSOC techniques significantly enhance existing state-of-the-art nominal TSC methods in ordinal problems. Notably, HC2 is overall outperformed by TSOC methodologies, particularly by O-MiniROCKET, including in terms of accuracy. 5) Contributing to the literature with the first baseline study for TSOC, which aims to encourage time series community for further enhancement in this novel field.}

The remainder of this paper is organised as follows: related works are described in \Cref{sec:prev_works}; \Cref{cap:tsoc_section} formalises preliminary definitions needed to present the different proposals; \Cref{cap:proposed_methods} defines the proposed TSOC methods; \Cref{cap:experiments} specifies the experimental setup and the datasets considered; finally, in \Cref{cap:conclusions}, we provide the conclusions and future research of our work.

\section{Related Works}\label{sec:prev_works}

\subsection{Time Series Classification (TSC)}

A range of TSC methodologies have been developed in recent years. The first taxonomy grouping these approaches by their typology was proposed in \cite{time_series_bakeoff}, and extended in \cite{bakeoff_redux}. This taxonomy categorises methodologies into eight distinct families: distance-based, interval-based, shapelets-based, feature-based, dictionary-based, convolutional-based, Deep Learning-based (DL), and hybrid methods. In this work, we focus on two of the best performing categories: convolutional-based and DL-based techniques. In the following, we provide a comprehensive introduction to both:

\noindent \textit{Convolutional-based} techniques: this family of methods tries to extract features from the input time series by applying a set of convolutional kernels. The RandOm Convolutional KErnel Transform (ROCKET) method \cite{rocket} was the first approach in this group of algorithms. It consists in training a Ridge regressor with features extracted from the application of kernels to the time series. In addition, two enhanced versions of this method were presented to the literature: MiniROCKET \cite{minirocket}, which proposes a computationally lighter kernel extraction process while maintaining a high accuracy, and MultiROCKET \cite{multirocket}, which extends the kernel extraction by applying it to the first order difference of the time series and performing more pooling operators. MultiROCKET is significantly better than MiniROCKET in accuracy, but it is computationally more expensive. A recently established method in the literature is the HYbrid Dictionary–Rocket Architecture (HYDRA) \cite{hydra}. HYDRA is a combination of dictionary and convolutional techniques that achieves a highly competitive performance in comparison with existing approaches. HYDRA can be used in combination with any ROCKET-based methodology. In the current state-of-the-art, MultiROCKET combined with HYDRA (HYDRA+MultiROCKET) stands as the best TSC methodology when accounting for both accuracy and training time. \rev{Additionally, several recent approaches utilise ROCKET as the foundational method. Uribarri \textit{et al.} \cite{uribarri2024detach} proposed Detach-ROCKET, which introduces a sequential feature detachment step in the pipeline to identify and prune non-essential features from the feature extraction phase in ROCKET-based models. In \cite{foumani2024improving}, the authors combined two position encodings within a transformer block alongside a convolutional layer, resulting in a novel convolutional approach that achieves competitive results compared to other state-of-the-art TSC methods.}

\noindent \textit{Deep learning-based} models: DL has also been \rev{applied to TSC}. Among \rev{the most commonly used architectures for neural networks are} the MultiLayer Perceptron (MLP) \cite{resnet} and the Convolutional Neural Network (CNN), which typically serve as a baseline architecture for DL-based TSC methods. Notably, InceptionTime \cite{inception} stands out as the leading methodology within this category, being an ensemble of $5$ Inception Networks, which, \rev{as with other approaches such as} ResNet \cite{resnet}, is based on residual networks. Various adaptations and extensions of InceptionTime have been explored in the literature. For instance, InceptionFCN combines the Fully Connected Network (FCN) with the InceptionTime architecture \cite{inception_tsc_survey}, while H-InceptionTime employs hand-crafted filters in the convolutional layers of the network \cite{hand_crafted_dl}. Recently, LITE has been introduced as a novel architecture featuring a light inception module with boosting techniques, using less than 3\% of InceptionTime's number of parameters \cite{ismail2023lite}. Recurrent Neural Networks (RNNs) is another popular type of architectures for DL. However, their application in TSC is relatively limited, primarily due to their design, which focuses on predicting an output for each timestamp in the time series \cite{langkvist2014review}. Long Short-Term Memory (LSTM)-based approaches have also been proposed for addressing the TSC task. For example, \cite{ma2022difference} introduced a difference-guided representation learning network that utilised LSTM to model the temporal dependencies and dynamic evolution of time series data. Additionally, transformer-based DL has gained \rev{in popularity for} TSC, often employing a straightforward encoder structure comprising attention and feed-forward layers. Notable among these approaches is AutoTransformer \cite{ren2022autotransformer}, which uses a neural architecture search algorithm to identify the most suitable network architecture before passing the output to a multi-headed attention block. Furthermore, graph neural networks \cite{bai2024haqjsk,li2024guest} have recently been developed, demonstrating their potential due to their ability to model inter-temporal/channel relationships, areas where other DL-based methods often face challenges \cite{jin2024survey}.

Despite the progress and strong performance achieved with existing methodologies in TSC, the performance achieved for the ordinal problems could be significantly improved with appropriate approaches. Following, we specifically focus on ordinal classifiers.

\subsection{Ordinal classification}

Over the last few years, several ordinal classification techniques based on nominal methodologies have been developed to exploit ordinality in the training process. In \cite{ordinal_svm}, a novel learning algorithm based on large margin rank boundaries was proposed for ordinal classification tasks, and the result can be seen as an adaptation of the Support Vector Machine (SVM) algorithm to ordinal problems. Subsequently, methods such as Support Vector Ordinal Regression with Explicit Constraints (SVOREX) or SVOR with Implicit Constraints (SVORIM) \cite{svm_3} have been proposed. Both methods are ordinal SVMs but with different constraints definition criteria in the optimisation process. In the same way, an extended kernel discriminant learning \cite{kernel_discriminant_analysis} algorithm has been proposed to work as an ordinal classifier by using a ranking constraint. This method is known as Kernel Discriminant Learning for Ordinal Regression (KDLOR) and has been proposed as an alternative to the ordinal SVM methods mentioned above. The key difference between these approaches is that SVOR methods do not take advantage of the global information of the data (especially SVOREX \rev{which} only considers adjacent classes when determining the thresholds) and also suffers from higher computational complexity. 

Other approaches to ordinal classification, such as \cite{gaussian_ordinal}, rely on Gaussian Processes (GP) to develop an ordinal regressor: GP for Ordinal Regression (GPOR). The performance of this new method has been compared against SVORs techniques in several datasets, and the results showed a good generalisation capacity and competitive performance. Furthermore, inspired by this approach and previous SVOR methods, Srijith et al. in \cite{gaussian_ordinal_2} have proposed a sparse GPOR using Leave-One-Out cross-validation to perform model selection (LOO-GPOR), a probabilistic least squares ordinal regressor (PLSOR) \cite{gaussian_ordinal_3}, and a semi-supervised ordinal regressor using GP \cite{gaussian_ordinal_4}. \rev{Several recent works have been published in the literature, such as ORFEO \cite{gomez2024orfeo}, which focuses on specific problems where the ordinal output is derived from the original continuous output. ORFEO is an artificial neural network that simultaneously optimises both outputs using a loss function that linearly combines ordinal classification and regression outputs. Marudi \textit{et al.} \cite{marudi2024decision} presented a novel decision tree-based method for ordinal classification that utilises a generalisation of the entropy measure for ordinal variables, along with an ordinal information gain ratio to assess the importance of each variable in the decision-making process.}

DL for ordinal classification problems is an area that has been less explored up to now, with some works such as \cite{clm} or \cite{clm_2}, which studied the application of a new output layer based on Cumulative Link Models (CLMs) together with a Quadratic Weighted Kappa (QWK) loss function. This approach has been proposed for tackling ordinal image classification problems, for which excellent results are achieved in comparison to nominal approaches. A recent work \cite{clm_novel} has taken this idea to approach the field of Aesthetic Quality Control (AQC), developing a DL architecture with CLM as the output layer and cross-entropy as the cost function. The results have demonstrated the capacity of the network to take advantage of the ordinal nature of AQC problems, improving performance over previous nominal approximations. \rev{Furthermore, Rosati \textit{et al.} \cite{rosati2024learning} proposed two novel ordinal-hierarchical deep learning methodologies: hierarchical Conditional Likelihood Models (CLM), based on the CLM framework, and hierarchical-ordinal binary decomposition. Both approaches effectively model the ordinal structure across different hierarchical levels of the labels. Finally, \cite{vargas2023generalised} introduced a soft labelling approach based on generalised triangular distributions, which enables the model to minimise errors associated in distant classes on the ordinal scale.} 

At this point, once both main fields tackled in this work have been introduced, Time Series Ordinal Classification (TSOC) can be defined. TSOC consists in the application of ordinal classifiers to ordinal time series problems, in such a way that the ordinal classifiers are able to consider and exploit the ordinal information present in the label space. TSOC is a recently established field only covered by a small set of conference papers \cite{shapelets,tsoc_2,tsoc_1}. The first two works have proposed an ordinal shapelet transformer in which the ordinal information of the data is exploited in two ways: 1) by introducing a novel shapelet quality measure in which ordinal information is taken into account, and 2) by using an ordinal classifier instead of a nominal one, specifically the Proportional Odds Model (POM) \cite{pom} and SVORIM \cite{svm_3} have been tested as final classifiers. Results have demonstrated that ordinal approaches significantly outperform nominal TSC approaches. Later, in \cite{tsoc_1}, the effect of using different $L_p$ norms for the computation of the shapelet quality has been studied. As can be seen, TSOC has room for significant improvement.

\section{Preliminary definitions}\label{cap:tsoc_section}
In this section, \rev{we provide formal definitions of TSOC concepts}. Moreover, we also define a family of probabilistic ordinal classifiers (\rev{known as CLMs}), which will be used for the TSOC methods proposed in this paper.

\subsection{Time series and related supervised learning tasks}\label{cap:ts_def}

In supervised learning time series problems, we are provided a training dataset including a set of labelled time series, $D=\{(\mathbf{t}_1,y_1),(\mathbf{t}_2,y_2),\ldots,(\mathbf{t}_N,y_N)\}$, where $N$ is the number of time series, $\mathbf{t}_i$ is the $i$-th time series, $y_i$ is the label assigned to it, and $i\in\{1,\ldots,N\}$. In general, the objective is to learn a mapping function (model) able to accurately predict the labels for the time series of the test set. Every time series is a set of $C$ ordered sets (also known as channels) of real values. Although the most common type of time series are univariate ($C=1$), there is an increasing interest in multivariate time series ($C>1$). A multivariate time series with $C$ channels is defined as $\mathbf{t}_i=(\mathbf{t}_i^1,\ldots,\mathbf{t}_i^C)$, where the $c$-th channel is denoted as $\mathbf{t}_i^c = (t^c_{i,1}, t^c_{i,2}, \ldots t^c_{i,T})$, $c \in \{1, \ldots, C\}$, and $T$ is the length of the time series. In this paper, we focus on time series with a constant spacing of observation times, and we consider datasets where all the time series are equal-length. 

Depending on the nature of the labels, the supervised time series problems can be categorised in Time Series Extrinsic Regression (TSER) \cite{tan2021time,guijo2024unsupervised}, nominal Time Series Classification (TSC) \cite{bakeoff_redux} or Time Series Ordinal Classification (TSOC). \rev{TSER covers problems} where the label to be predicted is a real number, i.e. $y_i\in\mathbb{R}$. For TSC, the label of each time series takes values in a discrete set of categories, $y_i\in\{\mathcal{C}_1,\mathcal{C}_2,\ldots, \mathcal{C}_q, \ldots, \mathcal{C}_Q\}$, where $q \in \{1, \ldots, Q\}$, and $Q$ is the number of classes of the problem. Finally, TSOC problems are TSC problems (with more than two classes) which include an order constraint in the labels.

\subsection{Time series ordinal classification (TSOC)}\label{cap:tsoc_def}

The focus of this paper is TSOC, where the categories show a natural order, in such a way that a relation $\mathcal{C}_1\prec\mathcal{C}_2\prec\ldots\prec\mathcal{C}_Q$ is present in label space. This relation is also present in TSER problems, where the $\prec$ operator is directly $<$ because we are working with real numbers instead of categories. However, the difference between TSOC and TSER is that the distance between categories is not known.

To provide a deeper understanding of this sort of problems, let us consider the example in \Cref{fig:ordinal_time_series_example}. In this case, the classification problem involves determining the age ranges using the outlines of the middle finger. The ranges exhibit a natural relationship and can be categorised into three levels: 0-6 years old, 7-12 years old, and 13-19 years old. TSOC problems involve two important characteristics: 1) misclassification costs are different depending on the error (\rev{an error that confuses the 0-6 years old class with the 13-19 years old class should be penalised more heavily than one that confuses the 0-6 years old class with the 7-12 years old class}); and 2) including the order information during learning should boost convergence.

\subsubsection{Cumulative Link Models (CLM) }\label{cap:clm_def}
CLM belong to a broad family of models known as threshold-based models, which are built on two main ideas: 1)  The existence of a real variable underlying the (discrete) output response, such that each class $\mathcal{C}_q$ belongs to a certain interval in $\mathbb{R}$. This hidden variable will be denoted as $y^*$ and is often referred to in the literature as \textit{latent variable}. 2) The use of a function $f$ that transforms a pattern of the input space $\mathbf{x} \in \mathbf{X} \subset \mathbb{R}^D $ (where $D$ is the input dimensionality) into the one-dimensional real space corresponding to the latent variable, $f: \mathbf{x} \in \mathbf{X} \subset \mathbb{R}^C \to \mathbb{R}$. The aim is to segment the real line into $Q$ consecutive ordered intervals. Each interval along the real line corresponds to a specific class $\mathcal{C}_q$. These intervals are defined by a threshold vector $\bm{\theta} = (\theta_1, \ldots, \theta_{Q-1})$, where $\bm{\theta} \in \mathbb{R}^{Q - 1}$. To guarantee that $P(y \preceq \mathcal{C}_q|\mathbf{x})$ increases with $q$, the thresholds vector must be non-decreasing, and, therefore, must satisfy the constraint $\theta_1 \leq \theta_2 \leq ... \leq \theta_{Q-1}$ \cite{pom}. Thus, $\theta_1$ is learned during the training process while the thresholds for $q=2$ to $q=Q-1$ can be obtained as $\theta_q = \theta_1 + \sum_{i=2}^q \gamma_i^2$. $\boldsymbol{\gamma} = (\gamma_2, \gamma_3, \ldots, \gamma_{Q-1})$ is also learned during the training process. In this way, the aforementioned constraint is always satisfied.  Considering this setting, an input $\mathbf{x}_i$ is associated with an output class $\mathcal{C}_q$ if $f(\mathbf{x}_i) \in [\theta_{q-1}, \theta_q]$. To completely cover the domain of the real line, $\theta_0$ and $\theta_Q$ are set to $-\infty$ and $+\infty$ respectively, and are not considered part of the vector $\bm{\theta}$.

In view of the foregoing, the special case of CLM \rev{considers} the latent variable model: $y^* = f(\mathbf{x}) + \epsilon$, where $\epsilon$ is the random error component with zero mean, $E(\epsilon)=0$, where $E$ represent the expected value, for which a certain distribution $F_\epsilon$ is assumed. Moreover, it assumes a linear latent variable $f( \mathbf{x}) = \mathbf{w}^T\mathbf{x}$. On that account, it is satisfied that:
\begin{equation}
\begin{aligned}
P(y \preceq \mathcal{C}_q|\mathbf{x}) & = P(y^* \leq \theta_q) \\ & = P(\mathbf{w}^T\mathbf{x} + \epsilon \leq \theta_q) \\ & = P(\epsilon \leq \theta_q - \mathbf{w}^T\mathbf{x}),
\end{aligned}
\label{error_component_probability}
\end{equation}taking $g$ as the probability distribution function (p.d.f) of the assumed $F_\epsilon$, we get to:
\begin{equation}
g(\theta_q - \mathbf{w}^T\mathbf{x}) = P(y \preceq \mathcal{C}_q|\mathbf{x}).
\label{function_g}
\end{equation}Therefore, the learning objective of the CLM is to find the set of optimal thresholds $\bm{\theta}^*$, as well as the optimal parameters $\mathbf{w^*}$ of the function $f$, in such a way that, given a pattern of the test set $(\mathbf{x}_i, y_i)$, the cumulative probabilities for each class $\mathcal{C}_q$, i.e. $P(y_i \preceq \mathcal{C}_q|\mathbf{x}_i)$, are as close as possible to the observed ones.

To this end, we need to define a loss function $\ell$ that represents the disagreement between the target label assigned to an input $\mathbf{x}_i$ and the output given by a prediction function $\text{pred}(\mathbf{p}_i)$, where $\mathbf{p}_i=(p_{i_1}, p_{i_2}, ..., p_{i_Q})$ is the vector of predicted probabilities, where $p_{i_q} = P(y_i = \mathcal{C}_q|\mathbf{x}_i) = P(y_i \preceq \mathcal{C}_q|\mathbf{x}_i) - P(y_i \preceq \mathcal{C}_{q-1}|\mathbf{x}_i)$. Hence, our objective is to find the function $\text{pred}$ that minimises the expected risk:
$$\mathcal{L}(\text{pred}) = \text{E}(\ell(y_i, \text{pred}(\mathbf{p}_i))),$$ where $\mathcal{L}$ is the risk function. There are different options for the expected risk within the context of ordinal \rev{classification} (e.g. the absolute difference in number of categories between the predicted and true labels). However, we cannot deal directly with this expected risk, mainly for two reasons: 1) the probability distribution that generates the patterns $(\mathbf{x_i}, y_i)$ of a dataset $D$ is unknown; 2) the function $\ell$ is naturally discontinuous since its two arguments belong to the discrete space defined by the output $y$. This, as stated in \cite{logistic_at}, can lead to an NP-hard problem. Therefore, the common practice is to approximate $\ell$ by the so called \textit{surrogate risk}, which, considering the CLM setting, can be defined as:
$$\mathcal{A}(f) = E(\psi(\bm{\theta}, f(\mathbf{x})),$$where $\psi : \mathbb{R}^{Q-1}\times\mathbb{R} \to \mathbb{R}$ is generally referred to as the \textit{surrogate loss} function. 
The need to introduce this surrogate terms is the main motivation for the existence of the previously introduced latent variable. As the values of $y^*$ are not a priori known, we work with the vector $\bm{\theta}$, so that, for a pair $(\mathbf{x}_i, y_i)$, where $y_i$ takes a value $\mathcal{C}_q$,  we want the output of $f(\mathbf{x}_i)$ to be as close as possible to the $\theta_q$ threshold. To ensure the ordering of $\bm{\theta}$ introduced above, we essentially have two options. \rev{The first is to generate its elements through an incremental function, such as the one employed in the CLM activation layer (see \Cref{sec:clm_activation_layer}). The second option is to design the cost function to naturally converge to a solution that maintains this ordering}, as demonstrated by the Logistic All-Threshold method \cite{logistic_at} used in our work.

\subsubsection{Logistic All-Threshold (LogisticAT)}
The LogisticAT method is a special case of CLM where the considered $F_\epsilon$ distribution is the logistic function.  This leads to an interesting property that characterises the so-called Proportional Odds Models (POM). This property states that the ratio of the odds for two patterns inputs  $\mathbf{x}_1$, $\mathbf{x}_2$ \cite{ordinal_survey} is: $$\frac{odds(y \preceq \mathcal{C}_q|\mathbf{x}_1)}{odds(y \preceq \mathcal{C}_q|\mathbf{x}_2)} = \text{exp}(-\mathbf{w}^T(\mathbf{x}_1 - \mathbf{x}_2)).$$ 

The main distinguishing feature of LogisticAT with respect to the POM is its (\textit{surrogate}) loss function, which is presented below:
\begin{equation}
\begin{aligned}
\psi_\mathrm{AT} = & \sum_{i=1}^N\left(\sum_{q=1}^{y_i-1}h(\theta_q - \mathbf{w}^T\mathbf{x}_i) + \sum_{q=y_i}^{Q-1}h(\mathbf{w}^T\mathbf{x}_i - \theta_q)\right) + \\ & \frac{\lambda}{2}\mathbf{w}^T\mathbf{w},        
\end{aligned}
\end{equation}
where $(\mathbf{x}_i, y_i)$ is a pattern of the training set defined in the form $h:\mathbb{R} \to \mathbb{R}$, $h(z) = \text{log}(1 + \text{exp}(z))$, $\mathbf{w}$ is the array of parameters associated with the function $f(\mathbf{x}) = \mathbf{w}^T\mathbf{x}$ presented above, and $\lambda$ is the regularisation term, which is adjusted by cross-validation in our experiments (see \Cref{cap:experiments}). Observe that in $\psi_{AT}$, by taking into account all the thresholds in $\bm{\theta}$ for each input $(\mathbf{x}_i, y_i)$, we ensure that at the point of convergence the ordering of $\bm{\theta}$ is satisfied.

This model is used for the ordinal convolutional-based techniques (O-ROCKET, O-Mini-Rocket, and O-MultiROCKET) discussed in \Cref{sec:convolutional-based}.

\subsubsection{CLM Activation Layer}\label{sec:clm_activation_layer}
The \rev{aforementioned} mathematical structure of the \rev{CLM} can be utilised as link function in DL architectures \cite{clm}. On this wise, the one dimensional output of a deep network is mapped to a set of predicted probabilities through a CLM, for which we considered (as in the case of the LogisticAT method) a logistic distribution for the error component $\epsilon$. In order to ensure the ordering of the vector $\bm{\theta}$, the next definition is considered: $\theta_q = \theta_1 + \sum_{i=1}^{q - 1}\alpha_i^2$, with $q \in \{2, ..., Q-1 \}$, then $\theta_1$ and $\alpha_2,\ldots,\alpha_{Q-1}$ can be learnt without constraints.

In addition, a cost function that takes greater account of ordinality is considered \cite{qwk_loss}. This function, denoted as $\psi_{\text{\tiny QWK}}$, is based in the Quadratic Weighted Kappa (QWK) metric, and it is defined in terms of the predicted probabilities:
\begin{equation}
\nonumber
\psi_{\text{\tiny QWK}} = \frac{\sum\limits_{i=1}^N \sum\limits_{q=1}^Q \omega_{y_i, q} P(y = \mathcal{C}_q | \mathbf{x}_i)}{\sum\limits_{q=1}^Q \frac{N_q}{N} \sum\limits_{j=1}^Q ( \omega_{j,q} \sum\limits_{i=1}^N P(y = \mathcal{C}_j | \mathbf{x}_i))},
\end{equation}
where $\psi_{\text{\tiny QWK}} \in [0,2]$, ($\mathbf{x}_i, y_i$) is the i-th sample of the data, $N_q$ is the number of patterns labeled with the $q$-th class, and $\omega_{j,q}$ are the elements of the penalization matrix, where $\omega_{j,q} = \frac{(j-q)^2}{(Q-1)^2}$.

This setting is implemented in the ordinal DL methodologies (O-InceptionTime, O-ResNet, O-LITETime, and O-CNN) discussed in \Cref{cap:deep_learning_methods}.

\section{Proposed TSOC methods}\label{cap:proposed_methods}
In this section, we introduce the seven TSOC methodologies. Within the category of Convolutional-based techniques (\Cref{sec:convolutional-based}), we present various adaptations to the ROCKET-family methods: O-ROCKET, O-MiniROCKET, and O-MultiROCKET. In addition, in line with DL techniques (\Cref{cap:deep_learning_methods}), we introduce the O-InceptionTime, O-ResNet, O-LITETime and O-CNN methodologies.

\subsection{Convolutional-based techniques}\label{sec:convolutional-based}
This family of TSC algorithms was \rev{firstly proposed} in \cite{rocket} with the development of the ROCKET method for TSC. The main idea is to extract features from the input time series by applying a set of convolutional kernels. This convolution process involves a sliding dot product that depends on the properties of the kernel being applied, which are: the values of the kernels or weights ($\mathbf{w}$), the length ($n$), which sets the kernel extension (or number of weights); dilation ($d$), which is a popular technique to increase the length of the kernel without increasing the length of the resulting convolution. This is done by ignoring one out of every two values of the time series, effectively adding empty cells in the kernels \cite{rocket_transform}; padding ($p$), which adds to the beginning and end of the series a vector filled with zeros to control where the middle weight of the first and last kernel to be applied falls; and bias ($b$), a real value that is added to the kernel convolution result.

In addition to the original version of ROCKET, the authors subsequently presented two improved versions: MiniROCKET \cite{minirocket} and MultiROCKET \cite{multirocket}. They all share the same architecture, comprising four sequentially applied phases: 1) kernel convolution transform, 2) pooling operations, 3) standard scaler, and 4) the final LogisticAT classifier.

\rev{We} propose the use of the LogisticAT method introduced in \Cref{cap:clm_def} as \rev{the} final classifier, with a built-in cross validation of the $\lambda$ regularisation parameter. With respect to the standardisation process and the final classifier, both remain unchanged. Conversely, the first two phases (kernel convolution and pooling operations), are different for each of the three alternatives. A comparison between the configuration of the three methods is presented in \Cref{tab:rocket-versions}.

\begin{table}[ht!]
\setlength{\tabcolsep}{3pt}
\renewcommand{\arraystretch}{1.35}
\begin{tabular}{llll}
\cline{1-4}
         & \textbf{O-ROCKET} & \textbf{O-MiniROCKET} & \textbf{O-MultiROCKET}                                                   \\ \hline
Kernel length & \{7, 9, 11\} & \{9\}           & \{9\}           \\
Weights       & $\mathcal{N}(0,1)$       & \{-1, 2\}   & \{-1, 2\}   \\
Dilation      & $\mathcal{U}(-1,1)$      & $\{2^0, \ldots, 2^{a}\}$ & $\{2^0, \ldots, 2^{a}\}$ \\
Use of Padding     & random       & alternative       & alternative       \\
\begin{tabular}[c]{@{}l@{}}Pooling \\ operations\end{tabular} & [PPV, GMP]       & [PPV]                  & \begin{tabular}[c]{@{}l@{}}[PPV, GMP, MPV,\\ MIPV, LSPV]\end{tabular} \\ 
Num. Features & 20,000 & 10,000 & 50,000 \\\hline
\end{tabular}
\caption{Differences between O-ROCKET approaches.}\label{tab:rocket-versions}
\end{table}

\subsubsection{O-ROCKET}\label{sec:rocket_subsection}
ROCKET is a significant contribution to the state-of-the-art in TSC, as it is capable of achieving excellent performance in a fraction of computational time. ROCKET applies to the input time series a large set of kernels generated with random properties: length $n$ is randomly sampled from $\{7, 9, 11\}$ with equal probability; weights are sampled from a normal distribution, $\forall w_k \in \mathbf{W}$,$w_k \sim \mathcal{N}(0, 1)$; bias $b$ is sampled from an uniform distribution, $b \sim \mathcal{U}(-1, 1)$; dilation $d = \lfloor 2^x\rfloor, x \sim \mathcal{U}(0, A), A = log_2 \frac{T - 1}{n - 1}$ where $T$ denotes the time series length, and $n$ the kernel length; and padding, which is denoted by $p$ and is applied or not with the same probability in each kernel. In the case it is applied, its value is computed with $p = ((n-1) * d)/2$.

Once the kernel convolution is finished, two real-valued features are extracted from each kernel. \rev{The first is derived from Global Max Pooling (GMP), which involves selecting the highest value. The second feature is} the Proportion of Positives Values (PPV), being $\text{PPV} = \frac{1}{n} \sum_{i=0}^{n-1}[z_i > 0]$, where $z_i$ is the output of a single kernel convolution. Hence, PPV is a real value ranging from zero to one that represent the percentage of kernel convolutions that are greater than zero. These extracted features are then fed as input to the LogisticAT classifier.


\subsubsection{O-MiniROCKET}
MiniROCKET \cite{minirocket} is a later version of ROCKET which is up to $75$ times faster than the original version, achieving similar performances. The main novelties with respect to ROCKET are the following, which make MiniROCKET become mostly (sometimes fully) deterministic:
\begin{itemize}
\item The kernel length $n$ is set to a fixed value of $9$.
\item Kernel weights $\mathbf{w}$ are restricted to two possible values, $-1$ and $2$.
\item A fixed dilation $d$ is computed according to the input length. The $d$ value can be in $\{2^0, \ldots, 2^{a}\}$, where $a = \text{log}_2(T-1)/(n-1)$.
\item Padding $p$ is applied or not alternatively for each kernel/dilation combination. This way, half of the kernels are computed with padding and the other half not. Whenever applied, its value is computed in the same way as in the O-ROCKET method.
\item Only the PPV aggregated feature is computed for each kernel.
\end{itemize}

\subsubsection{O-MultiROCKET}

This is the last version of the ROCKET family. In this novel variant, known as MultiROCKET \cite{multirocket}, two main adjustments are made to MiniROCKET: 1) the convolution is also done on the first order difference of the time series; and 2) three additional aggregated features are computed: the Mean Positives Values (MPV), Mean of Indices of Positives Values (MIPV) and the Longest Stretch of Positive Values (LSPV). These modifications substantially increase the computational expense of the algorithm but in turn achieves a significant increase in terms of accuracy.

\subsection{Deep learning techniques}\label{cap:deep_learning_methods}
We propose four types of DL architectures: the Ordinal ResNet (O-ResNet) methodology based on the residual network architecture \cite{resnet}, the Ordinal InceptionTime (O-InceptionTime) methodology based on the InceptionTime architecture \cite{inception}, the Ordinal Light Inception with boosTing tEchniques (O-LITETime) based on the LITETime model \cite{ismail2023lite}, and the Ordinal CNN (O-CNN) methodology, based on the CNN architecture \cite{zhao2017cnn}.

\subsubsection{O-ResNet}
\begin{figure}[ht]
	\centering
	\includegraphics[width=0.9\linewidth]{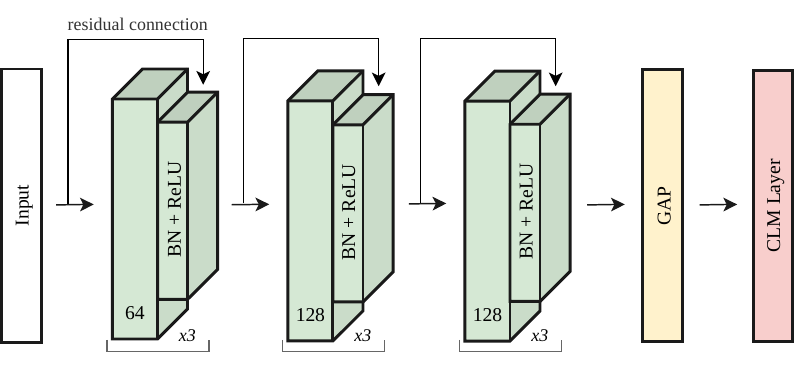}
	\caption{O-ResNet architecture. The notation \textit{x3} means that we have 3 stacked convolutional blocks inside each residual block. The final activation layer is the CLM (see \Cref{cap:clm_def}).}
	\label{fig:resnet_arch}
\end{figure}

The original residual architecture was proposed in \cite{original_resnet}. The main idea behind this sort of architectures \rev{was} to add shortcut connections between non-adjacent layers, which facilitates the flow of the gradient through the network. This work was extended to the TSC paradigm in \cite{resnet}, in which the input time series is passed sequentially through a set of three residually connected convolutional blocks, which, for ease of understanding, are referred to as super-blocks (SBlocks). The reason for this notation is given by the fact that each super-block $\text{SBlock}_{k}$ is composed of three simple convolutional blocks with $k$ filters each. In \Cref{fig:resnet_arch}, we refer to this fact with the \textit{x3} notation. In each $\text{SBlock}_{k}$, the convolution result is added to the original time series, and finally, a Batch Normalisation layer (BN) and a Rectified Linear Unit (ReLU) activation layer are applied. In our proposal, after the concatenated three SBlocks convolution, the final output of the network is obtained by applying a Global Average Pooling (GAP) layer followed by a CLM activation layer (see \Cref{fig:resnet_arch}). 


\subsubsection{O-InceptionTime}\label{cap:inceptiontime}
\begin{figure}[ht]
	\centering
	\includegraphics[width=0.9\linewidth]{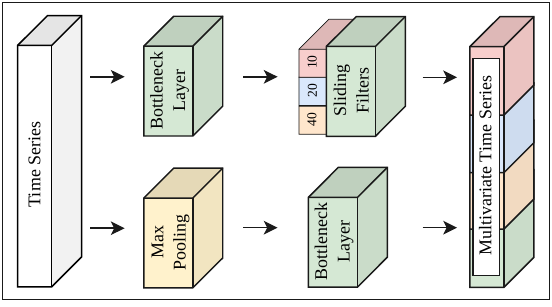}
	\caption{IM architecture. Two pipelines are applied in parallel to the input time series. On the first one, the bottleneck layer is applied, followed by $3$ sliding filter operations with sizes $40$, $20$ and $10$, respectively. On the second one, a Max Pooling is performed, followed by another bottleneck layer.}
	\label{fig:inception_module}
\end{figure}

\begin{figure}[ht]
	\centering
	\includegraphics[width=0.9\linewidth]{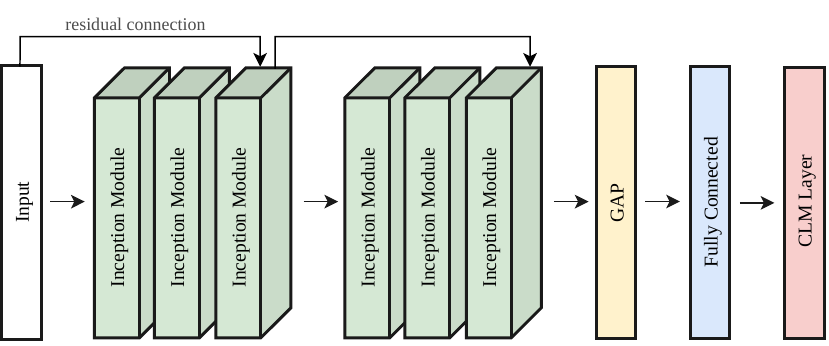}
	\caption{O-IN architecture, two blocks of three IMs with residual connections are stacked, the resulting multivariate time series is passed to a GAP layer, followed by a fully connected and a CLM activation layer (see \Cref{cap:clm_def}).}
	\label{fig:inceptiontime_arch}
\end{figure}

The O-InceptionTime architecture takes some of the main ideas of the ResNet architecture and the original InceptionTime methodology \cite{inception}, which includes concatenated convolution blocks together with residual connections between them. This method itself is an ensemble of $5$ O-Inception Networks (O-IN). An O-IN comprises two residual blocks, each consisting of three Inception Modules (IMs) that remain unchanged from the initial InceptionTime proposal (see \Cref{fig:inception_module}). O-IN applies a dimensionality reduction layer known as bottleneck, whose main goals are reducing the model complexity and avoiding overfitting. The convolution operation of a bottleneck layer consists in sliding $m$ filters of length $1$ with a stride equal to $1$. \Cref{fig:inceptiontime_arch} presents the O-IN architecture.

Specifically, an O-IN is composed of two residual blocks, each with 3 IMs. A GAP layer is applied to the output of the second block. Finally, the resulting features are fed to a fully connected layer, and finally, to the CLM activation layer.

As mentioned above, O-InceptionTime consists of $5$ O-INs initialised with random weights, where the final prediction is performed by a majority voting system. This ensemble is built with the purpose of reducing the variability inherent to the ResNet architectures \cite{resnet_variability}.
    
\subsubsection{O-LITETime}
This method, built upon the InceptionTime methodology, was developed to significantly reduce the computational complexity of its predecessor while maintaining competitive performance with other state-of-the-art methodologies. LITETime \cite{ismail2023lite} achieves this balance by possessing only 2.34\% of the number of parameters of InceptionTime. This reduction is primarily achieved through the use of bottleneck layers in the convolution process. To counteract this drastic simplification of the architecture, LITETime employs several boosting techniques: 1) \textit{Multiplexing}, which involves convolving the time series with kernels of different lengths simultaneously, allowing different convolution layers to be learned in parallel. 2) \textit{Dilation}, introduced in \Cref{sec:convolutional-based}, which helps to more easily capture long-term patterns in the time series. And, 3) \textit{custom filters}, i.e., handcrafted kernels that facilitate the learning of specific patterns (typically difficult to learn) in the time series. Finally, after the convolution process, the final classification step is performed using a CLM layer, specifically adapted to ordinal classification.

\subsubsection{O-CNN}
This approach is an adaptation of the standard CNN methodology for one-dimensional data in the form of time series \cite{zhao2017cnn}. The CNN method was a significant contribution as it introduced the use of convolutional networks to the field of TSC. This methodology alternates between applying convolution and pooling layers to the input time series. In the convolution layers, kernels of lengths 5, 7, and 9 are used. For the pooling layers, dimensionality reduction is achieved by segmenting the convolution output and calculating the average of each segment. Finally, the result of the convolution is used to feed a MLP layer, coupled with a CLM layer, that performs the final classification.

\section{Experimental settings and results}\label{cap:experiments}
First, we present the methods used for the comparison. Then we define the first version of the UCO TSOC repository, which contains a total of 29 datasets. After that, the experimental setup \rev{is described}. Finally, the results obtained by the seven proposed TSOC methods are compared with their nominal counterparts and with the rest of the nominal TSC approaches. Note that we provide open source \texttt{scikit-learn}-compatible implementations of all the methods benchmarked, a guide on how to reproduce experiments, and all the results obtained in an associated website\footnote{\label{note1}\url{https://uco.es/ayrna/grupos/tsoc-dl-conv}}. All ordinal approaches will be available in the \texttt{aeon} toolkit\footnote{\url{https://github.com/aeon-toolkit/aeon/}} \cite{aeon}, which has also been used for benchmarking and analysing the results obtained.

\subsection{Methodologies compared}\label{sec:comparisons}
To better understand how the proposed TSOC techniques perform, four different approaches have been run as a sanity check. The first two are a logistic regression \cite{logistic_regression_original} (LogReg) and a Extreme Gradient Boosting (XGBoost) \cite{xgboost} classifier. For these, time series are flattened into a vector concatenating all the channels. Thus, a multivariate time series with $C$ channels and length $T$ is transformed into a single vector of length $C \times T$. The third baseline method is the Time Series Forest (TSF) \cite{tsf}. Finally, the state-of-the-art approach in TSC, the HIVE-COTEv2 classifier (abbreviated as HC2) \cite{hc2}, has also been included to provide a deeper comparison. HC2 is an ensemble approach combining four approaches from different domains: convolutional-based, interval-based, dictionary-based, and shapelet-based. As explained above, these techniques are considered to enrich the experimentation, since they have been proven to be competitive in a wide range of problems.

\subsection{TSOC Datasets}\label{cap:tsoc_data}
Given that we are covering the TSOC paradigm, all of the datasets considered must be ordinal in nature. Therefore, we made a selection from two well-known repositories, the UCR TSC repository\footnote{\url{https://timeseriesclassification.com}}, from which we identified $9$ TSOC problems, and the Monash TSER repository\footnote{\url{http://tseregression.org/}}, from which $13$ additional datasets have been chosen. In the latter case, as these are regression datasets, the original continuous output variable has been discretised into five uniformly distributed bins. Furthermore, $5$ \rev{other} datasets have been built from historical price data from five important companies in the stock market, and other $2$ were \rev{formed} from data collected from the National Data Buoy Center (NDBC) \cite{gomez2022simultaneous}. Hence, a total of $29$ TSOC datasets have been considered in this work, with varying characteristics and backgrounds. This set of TSOC problems conforms the first version of the UCO TSOC repository\cref{note1}. Note that all the TSOC problems under consideration belong to the category of grouped continuous variables \cite{anderson_ordinal_variables}. This is attributed to the fact that all the TSOC problems, sourced from diverse data repositories, entail the discretisation of an underlying continuous variable. More information of the datasets, as well as specific information of the datasets such as the number of classes, time series length, number of channels and so on, is displayed in the Appendix A.

\subsection{Experimental settings}\label{cap:exp_setup}

The seven novel TSOC methodologies detailed in \Cref{cap:tsoc_section} as well as the baseline approaches have been applied to the whole UCO TSOC repository. The datasets in their original form are initially divided into train and test partitions. Due to the stochastic nature of the methodologies and to mitigate the risk of overfitting to the default training data partition, each experiment is repeated 30 times under varied conditions. Specifically, every experiment is conducted using a different seed for the initialisation of the methodologies and employing a distinct partition. For consistency, the first experiment is always executed with the default train/test partitions. Subsequent experiments, however, involve different partitions; the default train/test partitions are combined, and then 29 additional partitions are generated using a holdout procedure. Each of these partitions is created with a different seed (ranging from 1 to \rev{29}) while maintaining the same train/test proportion as the original dataset. We have to \rev{cross-validate} only one hyperparameter in our experiments, $\lambda$, which is the regularisation term of the LogisticAT classifier used in the ordinal convolutional-based methodologies. The set of $\lambda$ values to be tested are obtained according to $10^{-3 + \frac{6i}{9}}$, where $i \in \{0, 1, \ldots, 9\}$. A 5-fold stratified \rev{cross-validation} approach is used. The best $\lambda$ value is selected by Mean Absolute Error (MAE), as it best represents the performance of ordinal approaches.

As this work deals with ordinal classification, specific ordinal metrics should be considered. Concretely, MAE, 1-Off accuracy (1-OFF) \cite{chen2016cascaded} and Quadratic Weighted Kappa (QWK) \cite{cohen_function} are considered. All these metrics aim to quantify how close the predictions are to the actual variable on the ordinal scale. In addition, the Correct Classification Rate (CCR), widely used in nominal problems, is also computed.

\subsection{Results}
The first assessment is to determine whether the ordinal algorithms outperform their nominal counterparts. For this, the pairwise Wilcoxon signed-rank tests have been used with a significance level of $0.05$. Note that ranks have been averaged across all datasets before conducting the statistical tests. The results are presented in \Cref{tab:wilcoxon_pair_test}. Overall, TSOC methods are significantly better than nominal techniques. The exception is O-ResNet, where there is no discernible difference in any of the metrics.

\begin{table*}[ht]
\centering
\setlength{\tabcolsep}{5pt}
\renewcommand{\arraystretch}{1}
\begin{tabular}{cccccccc}
\toprule \toprule
 & \makecell{ROCKET vs \\ O-ROCKET} & \makecell{MiniROCKET vs \\ O-MiniROCKET} & \makecell{MultiROCKET vs \\ O-MultiROCKET} & \makecell{ResNet vs \\ O-ResNet} & \makecell{InceptionTime vs \\ O-InceptionTime} & \makecell{LITETime vs \\ O-LITETime} & \makecell{CNN vs \\ O-CNN}\\
 \cmidrule{2-8}
MAE & $\mathbf{< 0.01}$  & $\mathbf{< 0.01}$  & 0.156  & 0.442  & $\mathbf{< 0.01}$  & 0.305  & $\mathbf{< 0.01}$ \\
CCR & 0.053 & 0.086  & 0.304  & 0.831  & $\mathbf{< 0.01}$  & 1.000  & 0.139 \\
QWK & $\mathbf{< 0.01}$  & $\mathbf{< 0.01}$  & $\mathbf{< 0.01}$  & 0.381  & 0.345  & $\mathbf{< 0.01}$  & 0.203 \\
1-OFF & $\mathbf{< 0.01}$  & $\mathbf{< 0.01}$  & $\mathbf{< 0.01}$  & 0.155  & $\mathbf{< 0.01}$  & $\mathbf{0.048}$  & $\mathbf{< 0.01}$ \\
\bottomrule \bottomrule
\end{tabular}
\caption{$p$-values of the Wilcoxon paired tests comparing TSC versus TSOC versions of the methodologies proposed.}
\label{tab:wilcoxon_pair_test}
\end{table*}

\begin{figure}[ht]
    \begin{subfigure}[b]{0.45\textwidth}
    \includegraphics[width=\linewidth]{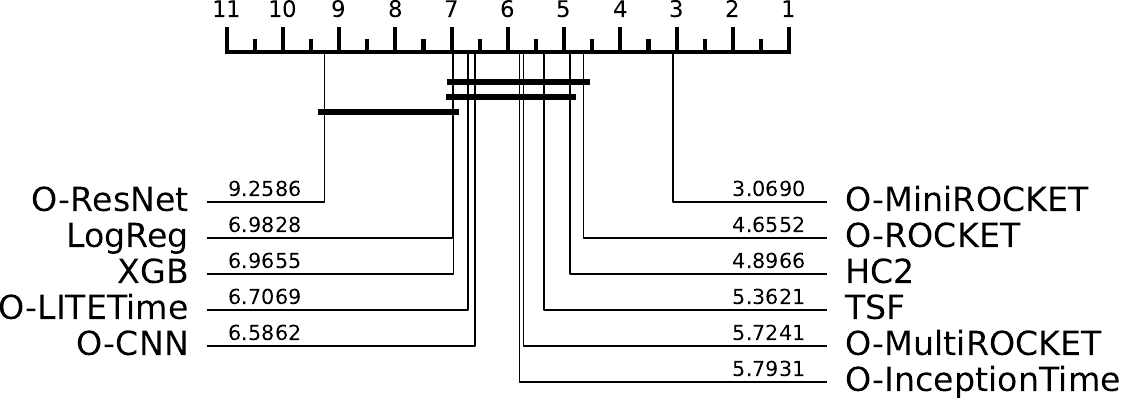}
   	\caption{Results in terms of MAE.}
   	\label{fig:cdd_all_mae}
    \end{subfigure}\\
    
    \begin{subfigure}[b]{0.45\textwidth}
    \includegraphics[width=\linewidth]{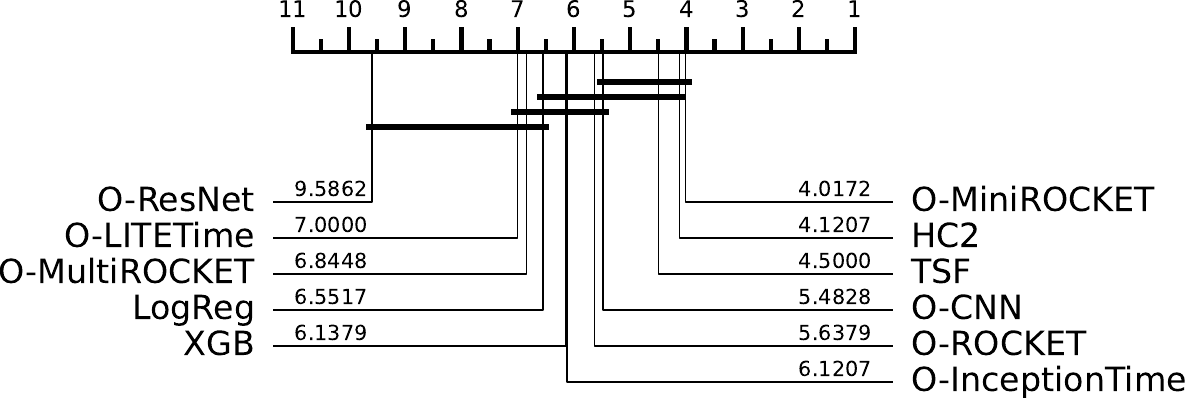}
   	\caption{Results in terms of CCR.}
   	\label{fig:cdd_all_ccr}
    \end{subfigure}\\
    
    \begin{subfigure}[b]{0.45\textwidth}
    \includegraphics[width=\linewidth]{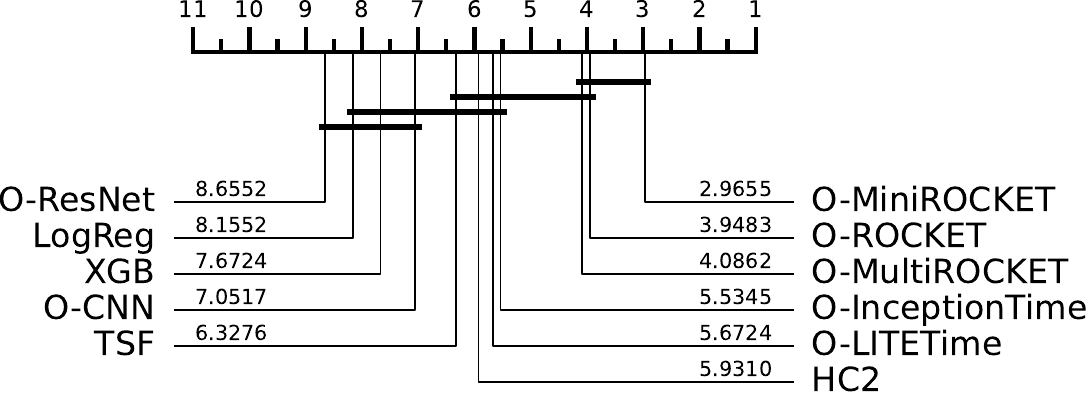}
   	\caption{Results in terms of 1-OFF.}
   	\label{fig:cdd_all_off1}
    \end{subfigure}\\
    
    \begin{subfigure}[b]{0.45\textwidth}
    \includegraphics[width=\linewidth]{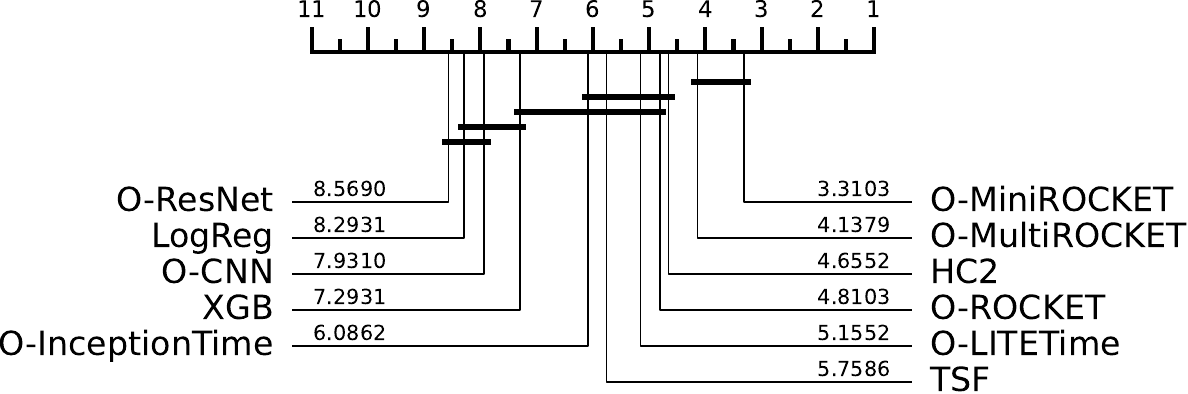}
   	\caption{Results in terms of QWK.}
   	\label{fig:cdd_all_qwk}
    \end{subfigure}
\caption{Comparison between the TSOC methodologies and the baseline approaches described in \Cref{cap:proposed_methods}. Methodologies are ordered based on the average rank over 30 resamples of train and test splits.}
\label{fig:nemenyi_global_test}
\end{figure}

The second experiment will determine which is the state-of-the-art approach in TSOC. Specifically, the seven ordinal versions of the proposed approaches are benchmarked along with the four comparative approaches introduced in \Cref{sec:comparisons}: XGBoost, LogReg, TSF, and HC2. For this, an adaptation of the critical difference diagram \cite{demsar06comparisons} has been used. The methodologies have been grouped into cliques, suggesting no significant difference in rank. These cliques were formed using the Holm correction for multiple testing as detailed in \cite{garcia08pairwise}, with a significance level of $0.05$. The results are graphically presented in \Cref{fig:nemenyi_global_test}, revealing that O-MiniROCKET consistently outperforms other methods across all metrics. Convolution-based methodologies, in general, demonstrate superior performance across all metrics. For MAE, O-MiniROCKET emerges as the top-performing approach, achieving significantly better results compared to all the other methods. Focusing now on the performance of the deep learners and keeping the convolution-based techniques aside, the O-InceptionTime is the best in terms of MAE and 1-OFF, being outperformed by O-CNN and by O-LITETime in terms of CCR and QWK, respectively. This is because deep learners require more training patterns to improve their performance. Notably, TSF and HC2 achieve their best performance for CCR, which aligns with their design objective. Moreover, except for MAE, where the O-MiniROCKET's superiority is statistically significant, and QWK, where both the O-MiniROCKET and O-MultiROCKET are the leading approaches (top clique), there are no significant differences among the top-performing methods for the remaining metrics. Therefore, it can be said that there is still room for better algorithms to be adapted and developed for TSOC problems. Furthermore, another way of improvement could be extending the problems archive, to include a wider range of datasets.

\begin{figure}[ht]
    \centering
	\includegraphics[width=0.9\linewidth]{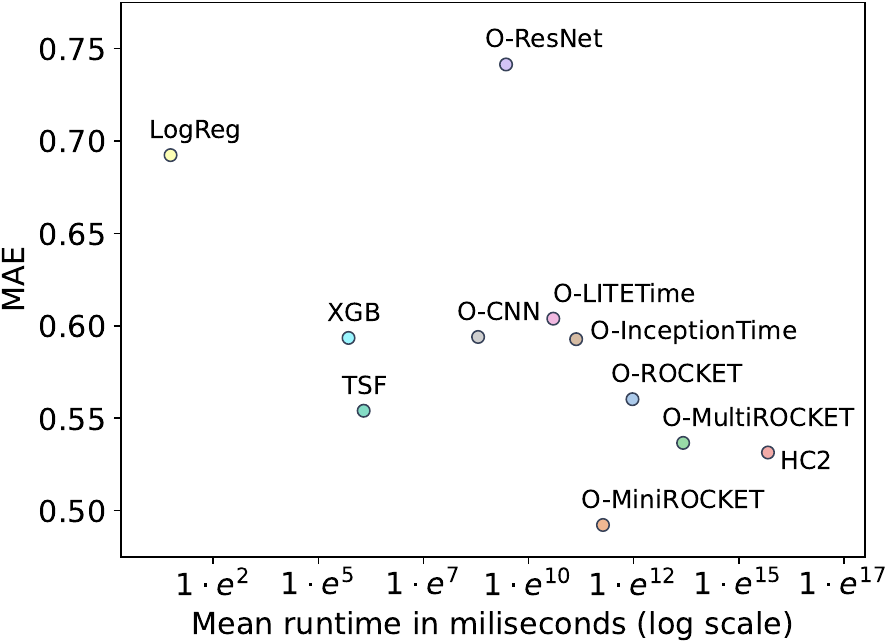}
    \caption{Run time in milliseconds (log scale average over all problems) plotted against mean MAE in the test sets.}
	\label{fig:runtimes}
\end{figure}

With the aim of evaluating the computational load of the methods, \Cref{fig:runtimes} compares the execution times in relation to the MAE. It is evident that ordinal ROCKET-based methodologies exhibit slower performance compared to other TSOC approaches such as O-ResNet or O-InceptionTime (run in GPU), but are faster than the HC2, the state-of-the-art nominal TSC approach. However, the O-ROCKET family methods outperform other methodologies, achieving over a 7\% improvement in mean MAE compared to HC2 and over 10\% against TSF, in the case of the O-MiniROCKET classifier. The increased computational load in the O-ROCKET-based methodologies is due to two main factors: 1) the extraction of kernels and computation of features, and 2) the application of the final ordinal classifier. The former can be examined by comparing the different versions of ROCKET-based methods. For example, O-MiniROCKET is the fastest as it employs only one pooling operation, while O-ROCKET and O-MultiROCKET are slower as they apply two and five pooling operations, respectively, as specified in \Cref{tab:rocket-versions}. The latter aspect represents a potential area for optimisation without compromising performance. Specifically, two potential enhancements for the ordinal versions of the O-ROCKET family methods include: 1) replacing the LogisticAT ordinal classifier with an ordinal version of the ridge classifier, which enables the computation of the projection matrix only once, or 2) leveraging \texttt{numba}\footnote{\url{https://numba.pydata.org/}}, a just-in-time compiler for \texttt{python}. Both enhancements could lead to a significant improvement in computational time. A special mention should be given to TSF, which demonstrates a favourable MAE-computational time trade-off, being orders of magnitude faster than ordinal O-ROCKET-based methodologies while achieving an acceptable mean MAE.

\begin{figure}[ht]
	\centering
	\includegraphics[width=0.9\linewidth]{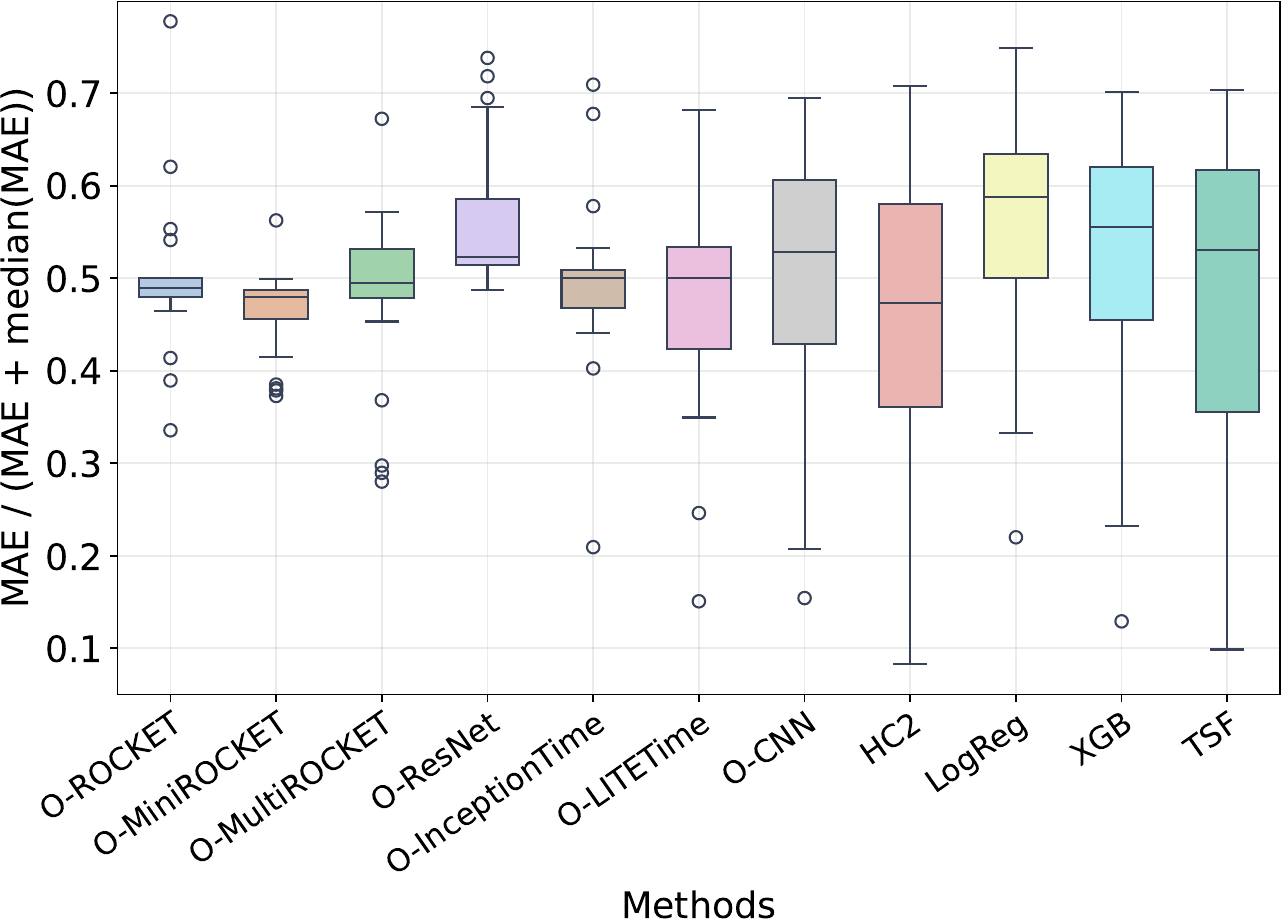}
    \caption{Boxplot of relative MAEs.}
	\label{fig:boxplots_relative}
\end{figure}

\begin{figure}[ht]
	\centering
	\includegraphics[width=0.9\linewidth]{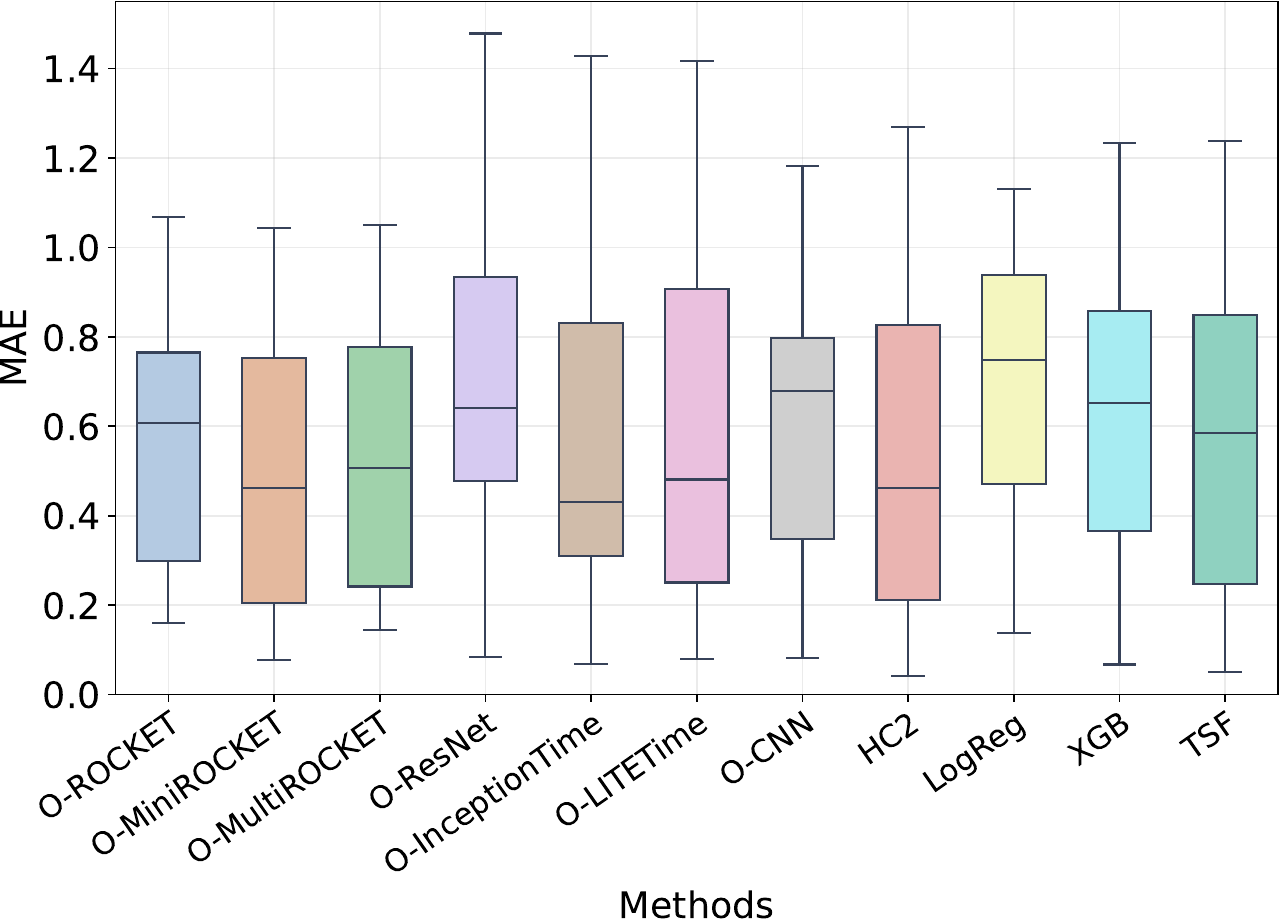}
    \caption{Boxplot of the results obtained in MAE.}
	\label{fig:boxplots_absolute}
\end{figure}

Additionally, the performance of each algorithm is compared in terms of relative MAE. The relative MAE is calculated by scaling the MAE of each approach with the median MAE for each dataset. \Cref{fig:boxplots_relative} presents the relative MAE for each method using boxplots. In this figure, values greater than $0.5$ indicate that the approach performs worse than the average method, while values smaller than $0.5$ indicate that the approach performs better than the average method. It can be observed that the O-ROCKET family exhibits a larger spread in the values, with most values being smaller than $0.5$. On the other hand, O-ResNet, O-CNN, the two standard approaches (LogReg and XGBoost), and the TSF approaches have values around $0.6$, indicating that they perform worse than the average algorithm. 
\rev{To complement the analysis presented in} \Cref{fig:boxplots_relative}, \Cref{fig:boxplots_absolute} \rev{displays standard boxplots illustrating the complete distribution of the results}. As observed, the majority of the results obtained by the ordinal convolutional approaches are consistently below a MAE of 1.0. The performance of the deep learning models is quite similar. Among the baseline nominal methodologies, HC2 stands out, demonstrating competitive performance relative to O-MiniROCKET. However, O-MiniROCKET consistently achieves the smallest Q1 and Q3 values and the smallest median value (on par with HC2), indicating its robustness and superior performance across the current TSOC repository.

\begin{figure*}[ht]
	\centering
	\includegraphics[width=0.9\linewidth]{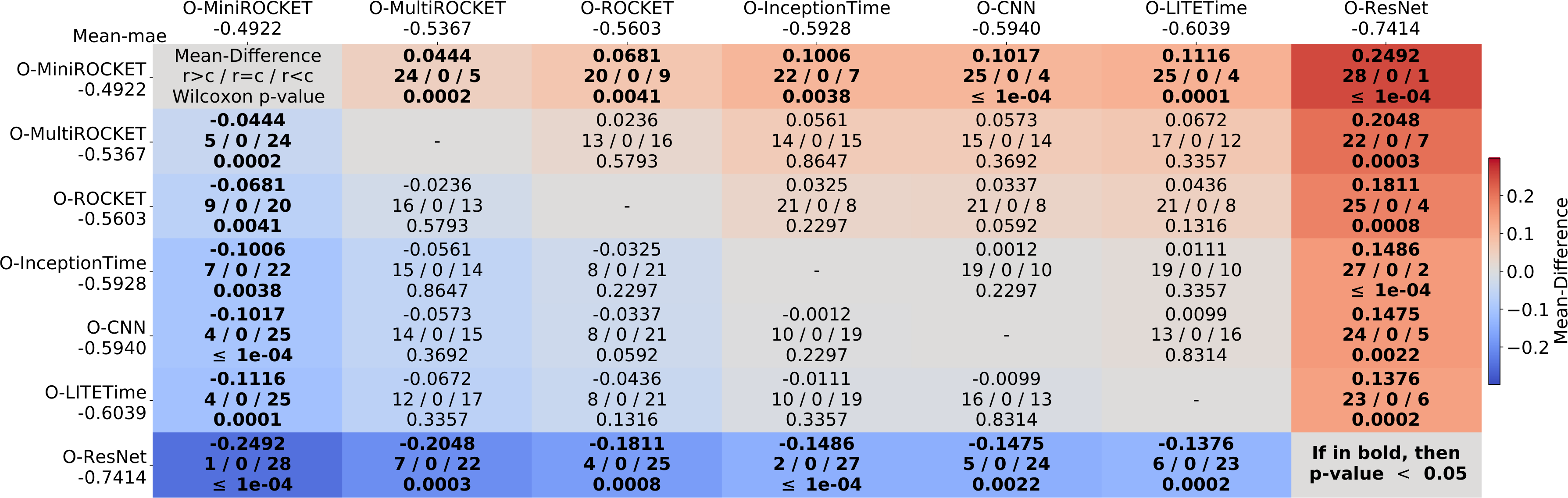}
    \caption{Multi-Comparison Matrix between ordinal methodologies in MAE. In each cell, three values are provided: 1) Average of differences in MAE. 2) Win/Ties/Losses counts. And 3) p-value of a Wilcoxon signed rank test.}
	\label{fig:heatmap_table}
\end{figure*}

Finally, in \Cref{fig:heatmap_table}, a global comparison in terms of MAE between the presented ordinal techniques is provided in the form of a full pairwise Multi-Comparison Matrix (MCM) \cite{mcm}. As can be observed, the MiniROCKET method is positioned as the best performing technique, obtaining a significant p-value ($<0.05$), against the rest of techniques.

Finally, to provide a deeper understanding of the characteristics of the datasets for which the different techniques are most suitable, the results have been analysed based on the number of classes, number of training patterns, time series length, and number of channels. This comparative analysis is presented in Appendix B due to space constraints.

\section{Conclusion}\label{cap:conclusions}

This paper presents the first \rev{application} of convolutional- and DL-based techniques to TSOC, \rev{to the best knowledge} of the authors. This area remains largely unexplored in comparison \rev{to} nominal TSC. \rev{One of our contributions} is the release of the UCO TSOC repository, including a total of 29 datasets \rev{from} various domains. We also contribute to the literature with ordinal versions of two main categories in TSC: convolutional (ROCKET, MiniROCKET and MultiROCKET) and DL-based approaches (InceptionTime, CNN, ResNet, and LITETime). The ordinal versions of these techniques have resulted in significant performance improvements on the selected ordinal datasets against the nominal ones. Specifically, O-MiniROCKET outperforms all other approaches across all performance metrics, with significant differences in MAE, and does so with an acceptable computational time. Remarkably, even in terms of CCR, a nominal performance metric, O-MiniROCKET outperforms HC2, the state-of-the-art approach in nominal TSC. Overall, O-InceptionTime stands out as the most promising DL model, significantly improving upon the results of other deep learners. All results, source code and guidance on reproducing experiments are available.

We recognise the potential for further advancements in TSOC algorithms. Incorporating \rev{alternative} TSC techniques \rev{offers} additional opportunities to explore this area and can serve as a baseline for future developments in this field. Another way of improvement is extending the experiments to develop ordinal counterparts of heterogeneous ensembles, such as HC2 \cite{hc2}. It may be possible to adapt and apply this approach to TSOC with careful consideration. Another potential research direction in TSOC is multi-objective optimisation, where two or more performance metrics may need to be simultaneously optimised, focusing on different aspects of ordinal classification \cite{gong2019multiobjective}.

Another area for improvement is in the UCO TSOC repository. Currently, the datasets are uniformly spaced, do not present missing values, and all time series are of equal length. However, we are conscious that many real-world TSOC datasets may not conform to these ideal conditions. Real-world TSOC problems often exhibit diverse characteristics, which may pose challenges such as irregular spacing, presence of missing values, or variable lengths of time series. We believe that as the TSOC field matures, there will be opportunities to develop novel methodologies capable of accepting this diversity in input data characteristics. We also would like to extend our repository with new problems. We greatly appreciate any contributions to this archive.

\ifCLASSOPTIONcompsoc
    \section*{Acknowledgments}
\else
    \section*{Acknowledgment}
\fi
The present study has been supported by the ``Agencia Estatal de Investigación (España)'' (grant ref.: PID2023-150663NB-C22 / AEI / 10.13039 / 501100011033), by the UK Engineering and Physical Sciences Research Council (grant ref.: EP/W030756/2), by the European Commission, AgriFoodTEF (grant ref.: DIGITAL-2022-CLOUD-AI-02, 101100622), by the Secretary of State for Digitalization and Artificial Intelligence ENIA International Chair (grant ref.: TSI-100921-2023-3), and by the University of Córdoba and Junta de Andalucía (grant ref.: PP2F\_L1\_15). R. Ayllón-Gavilán has been supported by the ``Instituto de Salud Carlos III'' (ISCIII) and EU (grant ref.: FI23/00163).

\ifCLASSOPTIONcaptionsoff
  \newpage
\fi

\bibliographystyle{IEEEtran}
\bibliography{IEEEabrv,bibliography}

\vspace{-40pt}
\begin{IEEEbiography}[{\includegraphics[width=1in,height=1.25in,clip,keepaspectratio]{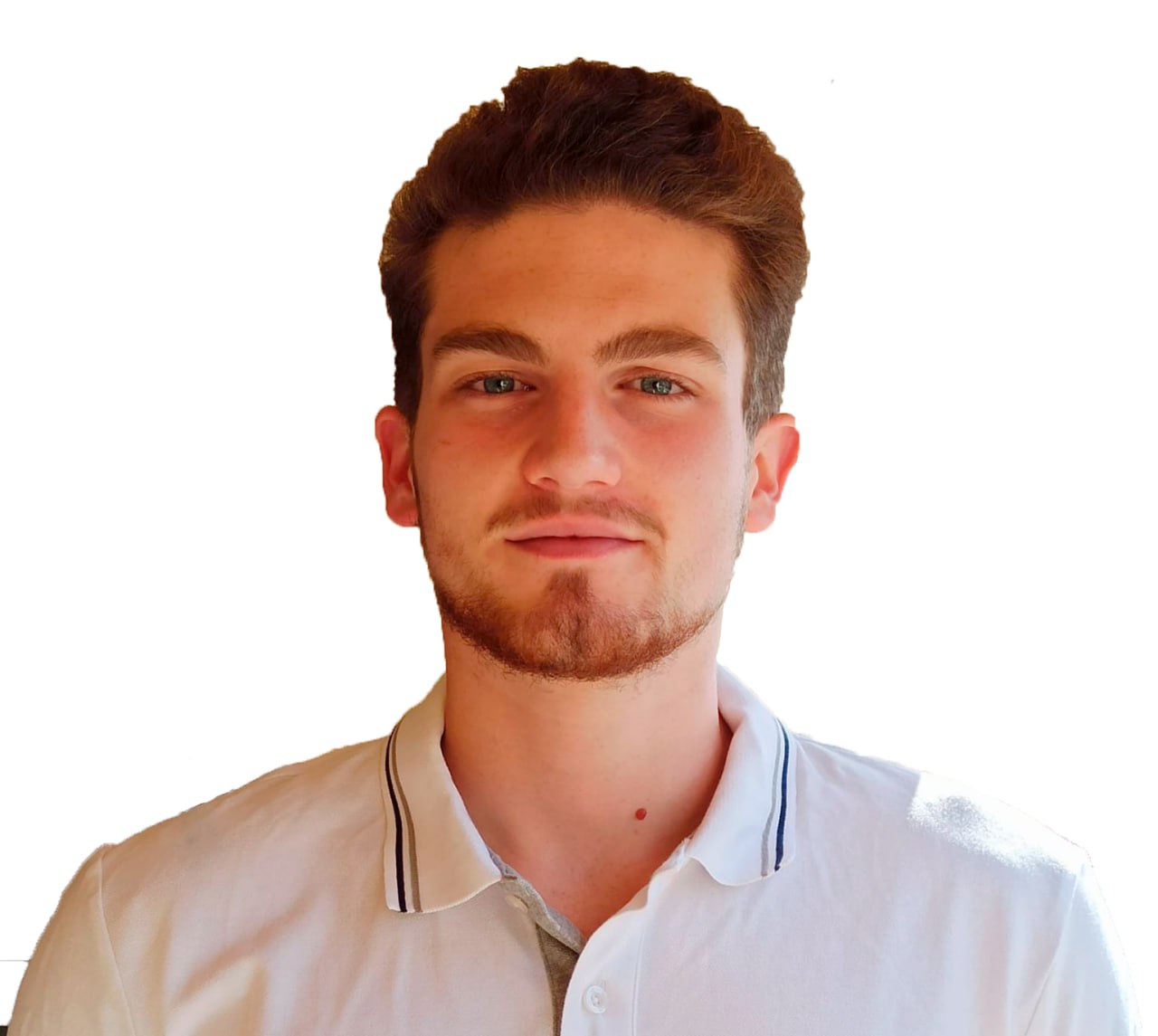}}]{\\Rafael Ayllón-Gavilán}
received the BSc degree in Computer Science in 2021 and the MSc degree in Artificial Intelligence in 2022. He is currently pursuing the PhD degree in Computer Science at the University of Córdoba. His current research concerns the time series classification domain, addressing the ordinal case together with its possible real-world applications.
\end{IEEEbiography}
\vspace{-40pt}
\begin{IEEEbiography}[{\includegraphics[width=1in,clip,keepaspectratio]{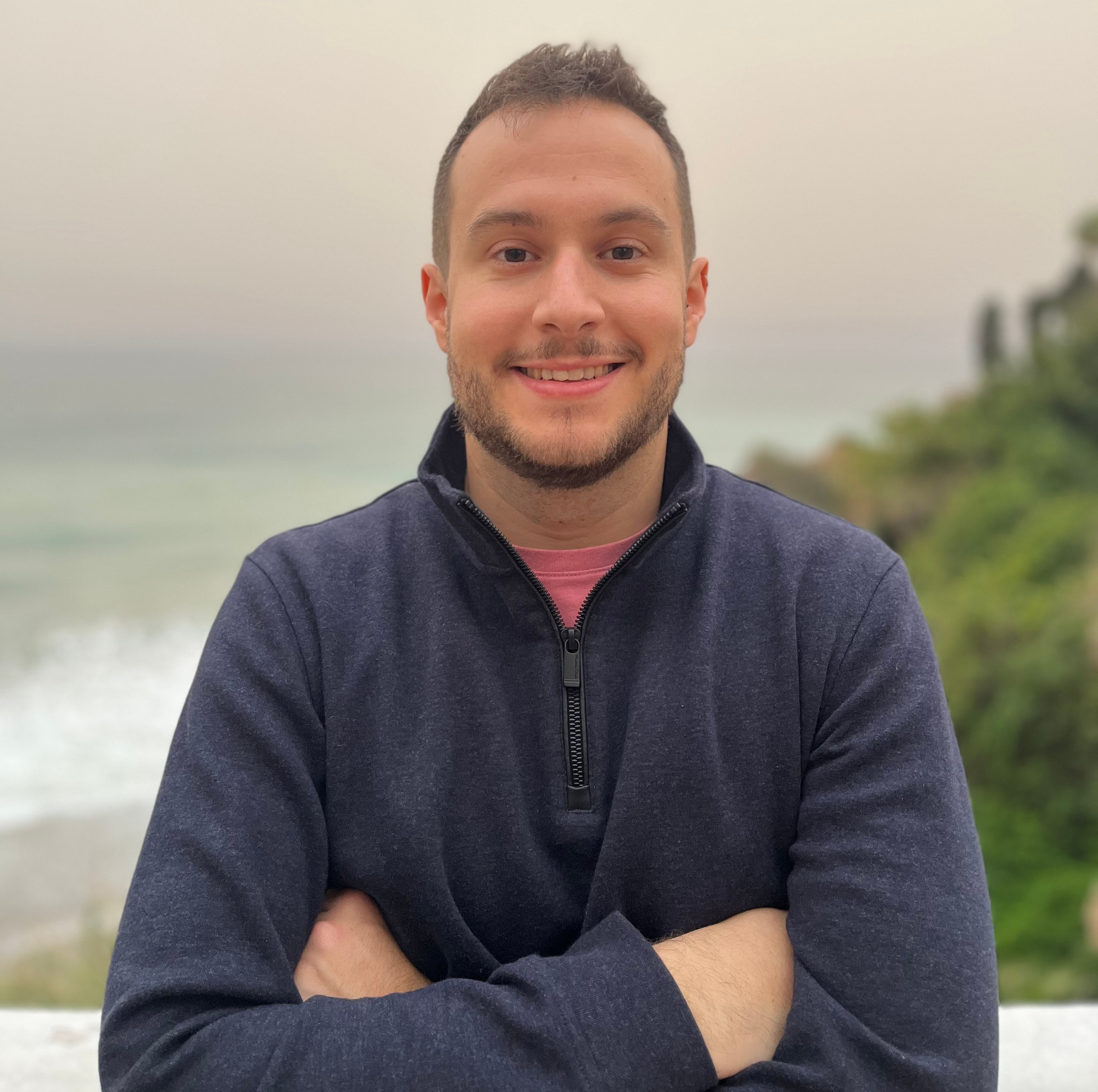}}]{\\David Guijo-Rubio}
(M'23) received the PhD degree in Computer Science from the University of Córdoba, Spain, in 2021. He is currently working as an Assistant Professor at the University of Córdoba. His current interests include different tasks applied to time series (classification --nominal and ordinal--, clustering, and regression). \end{IEEEbiography}
\vspace{-40pt}
\begin{IEEEbiography}[{\includegraphics[width=1in,height=1.5in,clip,keepaspectratio]{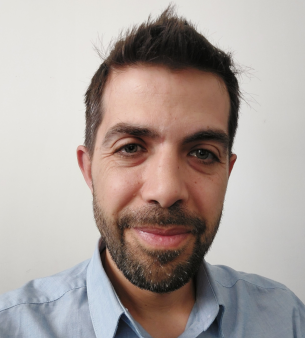}}]{\\Pedro A. Gutiérrez}
 (M'07-SM'15) received the PhD degree in Computer Science from the University of Granada, Spain, in 2009. He is currently a Full Professor at University of Córdoba. His research interests are supervised learning and ordinal classification. He serves on the Editorial board of the IEEE Transaction on Neural Networks and Learning Systems.
\end{IEEEbiography}
\vspace{-40pt}
\begin{IEEEbiography}
[{\includegraphics[width=1in,height=2in,clip,keepaspectratio]{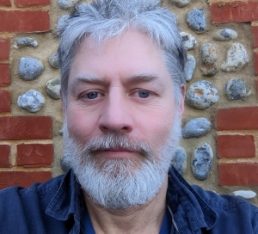}}]{\\Anthony Bagnall}
received the PhD degree in Computer Science from the University of East Anglia, in 2001. He is currently a Full Professor at University of Southampton, UK. His primary research interest is in time series machine learning, with focus on classification, but more recently looking at clustering and regression. He has a side interest in ensemble design. 
\end{IEEEbiography}
\vspace{-40pt}
\begin{IEEEbiography}[{\includegraphics[width=1in,trim={0cm 7cm 0cm 2.5cm},clip]{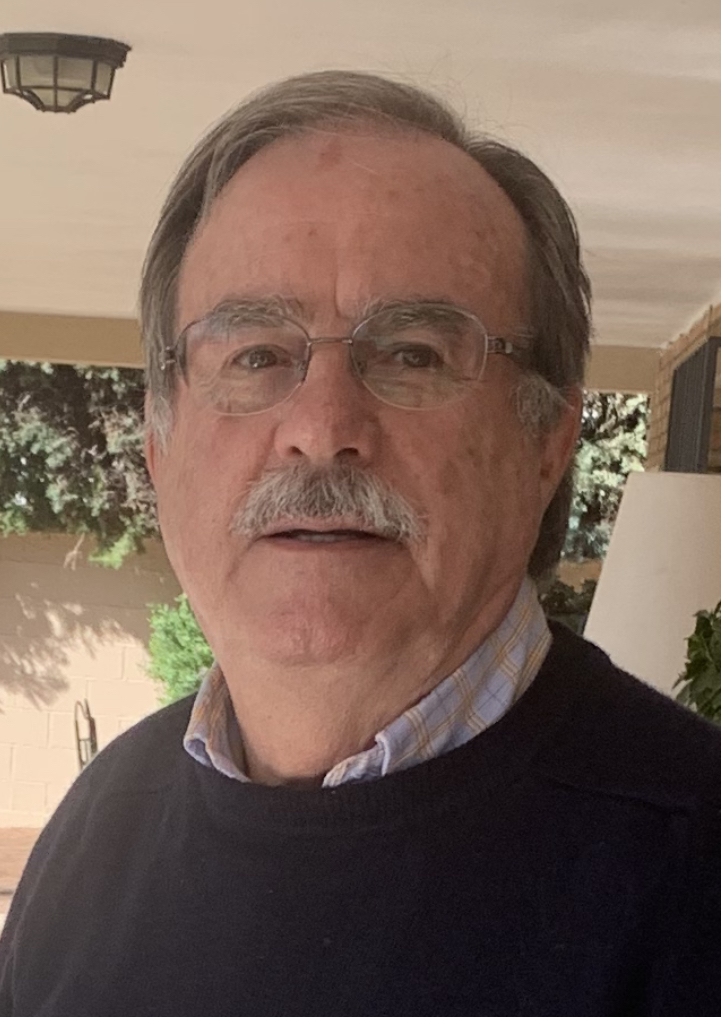}}]{\\César Hervás-Martínez}
(M'07-SM'14) received the PhD in Ma\-the\-ma\-tics from the University of Seville, Spain, in 1986. He is currently a Full Professor of Computer Science at University of Córdoba, Spain, leading the AYRNA research group. His current research interests include neural networks, evolutionary computation, and the modelling of natural systems.
\end{IEEEbiography}
\vspace{-40pt}
\vfill

\end{document}


\title{Convolutional and Deep Learning based techniques for Time Series Ordinal Classification}

\author{Rafael Ayllón-Gavilán,
        David Guijo-Rubio$^*$, \textit{Member, IEEE},
        Pedro Antonio Gutiérrez, \textit{Senior Member, IEEE},
        Anthony Bagnall,
        César Hervás-Martínez, \textit{Senior Member, IEEE},
\IEEEcompsocitemizethanks{

\IEEEcompsocthanksitem $^*$: Corresponding author.\protect\\
\IEEEcompsocthanksitem R. Ayllón-Gavilán is with the Dept. of Clinical-Epidemiological Research in Primary Care, IMIBIC, Spain. E-mail: rafael.ayllon@imibic.org.\protect\\
\IEEEcompsocthanksitem D. Guijo-Rubio, P.A. Gutiérrez and C. Hervás-Martínez are with the Dept. of Computer Science and Numerical Analysis, University of Cordoba, Spain. E-mail: \{dguijo,pagutierrez,chervas\}@uco.es.\protect\\
\IEEEcompsocthanksitem A. Bagnall is with the Dept. of Electronics and Computer Science, University of Southampton, United Kingdom. E-mail: A.J.Bagnall@soton.ac.uk.}
\thanks{Manuscript received June 18, 2023; revised July 13, 2024.}}

\markboth{IEEE Transactions on Cybernetics}%
{Ayllón-Gavilán \MakeLowercase{\textit{et al.}}: Convolutional and Deep Learning based techniques for TSOC}

\IEEEtitleabstractindextext{%
\begin{abstract}
Time Series Classification (TSC) covers the supervised learning problem where input data is provided in the form of series of values observed through repeated measurements over time, and whose objective is to predict the category to which they belong. When the class values are ordinal, classifiers that take this into account can perform better than nominal classifiers. Time Series Ordinal Classification (TSOC) is the field covering this gap, yet unexplored in the literature. There are a wide range of time series problems showing an ordered label structure, and TSC techniques that ignore the order relationship discard useful information. Hence, this paper presents a first benchmarking of TSOC methodologies, exploiting the ordering of the target labels to boost the performance of current TSC state-of-the-art. Both convolutional- and deep learning-based methodologies (among the best performing alternatives for nominal TSC) are adapted for TSOC. For the experiments, a selection of $29$ ordinal problems from two well-known archives has been made. In this way, this paper contributes to the establishment of the state-of-the-art in TSOC. The results obtained by ordinal versions are found to be significantly better than current nominal TSC techniques in terms of ordinal performance metrics, outlining the importance of considering the ordering of the labels when dealing with this kind of problems.
\end{abstract}

\begin{IEEEkeywords}
time series machine learning, time series analysis, time series classification, ordinal classification
\end{IEEEkeywords}}

\maketitle

\IEEEdisplaynontitleabstractindextext

\IEEEpeerreviewmaketitle

\begin{appendices} 

\section{Data description} \label{sec:data}
First of all, we made a dataset selection from two time series repositories, the UCR TSC archive\footnote{\url{https://timeseriesclassification.com}}, and the Monash TSER archive\footnote{\url{http://tseregression.org/}}. For the TSC repository, we analysed the output variable of each problem to select those with an ordinal nature. We have identified seven univariate, and two multivariate datasets. Moreover, we have chosen \rev{13} additional datasets from the TSER archive. Given that it is a regression repository, we have discretised the original continuous output variable, grouping it into five uniformly distributed bins. Even though there are more regression datasets, we could only select these \rev{13} as it is difficult to achieve a reasonable class distribution for the rest (the output values are too close to obtain more than three representative classes). Furthermore, the datasets with unequal length time series and missing values are also discarded. Other five TSOC problems built from historical price data from five of the most important companies in the stock market are considered: Apple (AAPL), Amazon (AMZN), Google (GOOG), META and Microsoft (MSFT), and other two were built from data collected from the National Data Buoy Center (NDBC).

Data for the five TSOC problems built from historical price data has been collected from Yahoo Finance\footnote{\url{https://es.finance.yahoo.com/}} website, extracting weekly price data from each stock earliest available date to March 2023. The time series of each stock data are formed by its price returns over 53 weeks (around one year) before a varying date $t$. The output ordinal label corresponds to the price return in $t$. This value is discretised according to a set of predefined symmetrical thresholds $(-\infty, -0.05, -0.02, 0.02, 0.05, +\infty)$. Finally, two more TSOC problems were built from data collected from the National Data Buoy Center (NDBC)\footnote{\url{https://www.ndbc.noaa.gov/}}. Input multivariate time series are seven climatological variables collected from a reanalysis model, as detailed in \cite{gomez2022simultaneous}. Some of these variables are the air temperature, the pressure, or the relative humidity. Each time series comprises four values per day during four weeks, i.e. the length of the time series is $112$. The output label indicates the energy fluctuation level during the data collection period, with $0$ being the minimum and $3$ the highest level. The second problem also comprises multivariate time series. Same seven climatological variables are considered for this problem. This time, the length of the time series is $28$, as collects values during only one week (four values per day). The purpose is to estimate the wave height level during that period of time, from $0$ (the lowest height) to $3$ (the highest height). Note that the outputs of both datasets have been discretised in bins with identical widths.

\Cref{table:dataset_info} presents the datasets information used in this study.

\begin{table}[ht]
\setlength{\tabcolsep}{3pt}
\renewcommand{\arraystretch}{1.35}
\centering
\begin{tabular}{clccccc}
\toprule\toprule
& Dataset & \#Train & \#Test & $Q$ & $T$ & $C$ \\ \midrule
\multirow{9}{*}{\rotatebox[origin=c]{90}{TSC}}& AtrialFibrillation             &     15 &    15 &        3 &     640 & 2\\
& DistalPhalanxOAG      &    400 &   139 &        3 &      80 & 1 \\
& DistalPhalanxTW                &    400 &   139 &        6 &      80 & 1 \\
& EthanolConcentration           &    261 &   263 &        4 &    1751 & 3 \\
& EthanolLevel                   &    504 &   500 &        4 &    1751 & 1\\
& MiddlePhalanxOAG   &    400 &   154 &        3 &      80 & 1 \\
& MiddlePhalanxTW                &    399 &   154 &        6 &      80 & 1 \\
& ProximalPhalanxOAG &    400 &   205 &        3 &      80 & 1 \\
& ProximalPhalanxTW              &    400 &   205 &        6 &      80 & 1 \\ \midrule
\multirow{12}{*}{\rotatebox[origin=c]{90}{TSER}} & AcousticContaminationMadrid & 166 & 72 & 5 & 365 & 1 \\
& AppliancesEnergy            &     95 &    42 &        5 &     144 & 24\\
& BarCrawl6min & 140 & 61 & 5 & 360 & 3 \\
& Covid3Month                 &    140 & 61 & 5 & 84 & 1 \\ 
& DhakaHourlyAirQuality & 1447 & 621 & 5 & 24 & 1 \\
& GasSensorArrayAcetone & 324 & 140 & 3 & 7500 & 1 \\
& GasSensorArrayEthanol & 324 & 140 & 3 & 7500 & 1 \\
& NaturalGasPricesSentiment & 65 & 28 & 5 & 20 & 1\\
& ParkingBirmingham & 1391 & 597 & 5 & 14 & 1\\
& SolarRadiationAndalusia & 470 & 202 & 5 & 365 & 2\\
& TetuanEnergyConsumption & 254 & 110 & 5 & 144 & 5\\
& WaveDataTension & 1325 & 568 & 5 & 57 & 1\\
& WindTurbinePower & 596 & 256 & 5 & 144 & 1\\ \midrule
\multirow{7}{*}{\rotatebox[origin=c]{90}{new TSOC}} & AAPL                           &   1720 &   431 &        5 &      53 & 1 \\
& AMZN                           &   1035 &   259 &        5 &      53 & 1 \\
& GOOG                           &    732 &   183 &        5 &      53 & 1 \\
& META                           &    408 &   103 &        5 &      53 & 1 \\
& MSFT                           &   1501 &   376 &        5 &      53 & 1 \\
& USASouthwestEnergyFlux      &    327 &   141 &        4 &     112 & 7 \\
& USASouthwestSWH             &   1310 &   562 &        4 &      28 & 7 \\ \bottomrule \bottomrule
\end{tabular}
\caption{\rev{Characteristics of the datasets used in this work. \#Train stands for the number of training patterns, \#Test that of the test set, $Q$ the number of classes, $T$ for the time series length, and $C$ is the number of channels.}}
\label{table:dataset_info}
\end{table}

\section{TSOC methods comparison tables} \label{sec:data}
\rev{To further analyse the performance of TSOC approaches based on the main characteristics of the datasets, we have divided the performance results to gain insights into the characteristics under which different algorithms perform well. These analyses have been conducted using the average rank in terms of MAE. \Cref{tab:analysis_length} shows the differences when grouping by time series length. As observed, O-MiniROCKET performs better with shorter series (length $<$ 1000), while O-MultiROCKET is better suited for longer time series. \Cref{tab:analysis_classes} presents the average rank when the datasets are grouped by the number of classes. Overall, O-MiniROCKET is the best approach in all groups except for the 4-class datasets, where O-LITETime performs better. This is likely because this deep learner performs particularly well on the Ethanol datasets. \Cref{tab:analysis_trainpatterns} includes the analysis when grouping by the number of instances in the training sets. Here, O-MiniROCKET always outperforms others with a smaller number of instances. O-InceptionTime is consistently the second best with larger training set sizes. Finally, \Cref{tab:analysis_channels} compares the ranks when the datasets are grouped by the number of channels. In this case, O-MiniROCKET also achieves the best for all the categories.}

\begin{table}
\revfloat
\centering
\normalsize
\resizebox{0.5\textwidth}{!}{
\begin{tabular}{l@{\hskip 0.5cm}*{4}{c@{\hskip 0.125cm}}}
\toprule
\toprule
Method & 0--75 (10) & 75--150 (11) & 150--1000 (4) & $>$1000 (4) \\
\midrule
O-ROCKET & $\mathit{3.300}$ & $\mathit{2.727}$ & $\mathit{3.250}$ & $3.500$ \\
O-MiniROCKET & $\mathbf{2.000}$ & $\mathbf{1.727}$ & $\mathbf{2.000}$ & $\mathit{3.000}$ \\
O-MultiROCKET & $4.000$ & $4.818$ & $3.750$ & $\mathbf{2.250}$ \\
O-ResNet & $6.100$ & $5.818$ & $6.500$ & $6.750$ \\
O-InceptionTime & $3.400$ & $4.182$ & $3.500$ & $3.500$ \\
O-LITETime & $4.200$ & $4.818$ & $5.000$ & $3.750$ \\
O-CNN & $5.000$ & $3.909$ & $4.000$ & $5.250$ \\
\bottomrule \bottomrule
\end{tabular}
}
\caption{\rev{Average rank MAE split by time series length.}}
\label{tab:analysis_length}
\end{table}

\begin{table}
\revfloat
\centering
\normalsize
\begin{tabular}{l@{\hskip 0.6cm}*{4}{c@{\hskip 0.3cm}}}
\toprule
\toprule
Method & 3 (6) & 4 (4) & 5 (16) & 6 (3) \\
\midrule
O-ROCKET & $\mathit{2.500}$ & $3.750$ & $\mathit{3.438}$ & $\mathit{1.667}$ \\
O-MiniROCKET & $\mathbf{1.667}$ & $\mathit{2.750}$ & $\mathbf{2.125}$ & $\mathbf{1.333}$ \\
O-MultiROCKET & $4.000$ & $3.750$ & $3.875$ & $5.333$ \\
O-ResNet & $6.000$ & $6.500$ & $6.125$ & $6.000$ \\
O-InceptionTime & $4.333$ & $3.000$ & $3.562$ & $4.333$ \\
O-LITETime & $5.667$ & $\mathbf{2.500}$ & $4.438$ & $5.000$ \\
O-CNN & $3.833$ & $5.750$ & $4.438$ & $4.333$ \\
\bottomrule \bottomrule
\end{tabular}

\caption{\rev{Average rank MAE split by number of classes.}}
\label{tab:analysis_classes}
\end{table}

\begin{table}
\revfloat
\centering
\normalsize
\resizebox{0.5\textwidth}{!}{
\begin{tabular}{l@{\hskip 0.45cm}*{4}{c@{\hskip 0.25cm}}}
\toprule\toprule
Method & 0--200 (6) & 201--400 (11) & 401--800 (5) & $>$800 (7) \\
\midrule
O-ROCKET & $\mathit{3.000}$ & $\mathit{2.636}$ & $3.800$ & $3.429$ \\
O-MiniROCKET & $\mathbf{2.167}$ & $\mathbf{1.818}$ & $\mathbf{2.200}$ & $\mathbf{2.143}$ \\
O-MultiROCKET & $\mathit{3.000}$ & $4.455$ & $4.600$ & $3.857$ \\
O-ResNet & $6.500$ & $5.909$ & $6.800$ & $5.714$ \\
O-InceptionTime & $4.833$ & $3.818$ & $\mathit{2.800}$ & $\mathit{3.286}$ \\
O-LITETime & $5.333$ & $4.636$ & $3.800$ & $4.000$ \\
O-CNN & $3.167$ & $4.727$ & $4.000$ & $5.571$ \\
\bottomrule \bottomrule
\end{tabular}
}
\caption{\rev{Average rank MAE split by number of train patterns.}}
\label{tab:analysis_trainpatterns}
\end{table}

\begin{table}
\revfloat
\centering
\normalsize
\begin{tabular}{l@{\hskip 0.35cm}*{3}{c@{\hskip 0.2cm}}}
\toprule\toprule
Method & 1 (21) & 2--4 (4) & $>$4 (4) \\
\midrule
O-ROCKET & $\mathit{3.095}$ & $\mathit{2.750}$ & $3.500$ \\
O-MiniROCKET & $\mathbf{1.952}$ & $\mathbf{2.500}$ & $\mathbf{2.000}$ \\
O-MultiROCKET & $4.143$ & $3.000$ & $4.500$ \\
O-ResNet & $6.048$ & $6.500$ & $6.250$ \\
O-InceptionTime & $3.714$ & $3.750$ & $3.750$ \\
O-LITETime & $4.619$ & $5.500$ & $\mathit{2.750}$ \\
O-CNN & $4.429$ & $4.000$ & $5.250$ \\
\bottomrule \bottomrule
\end{tabular}
\caption{\rev{Average rank MAE split by number of channels.}}
\label{tab:analysis_channels}
\end{table}

\end{appendices}

\bibliographystyle{IEEEtran}
\bibliography{IEEEabrv,bibliography}

\vfill